\pdfoutput=1
\documentclass{bmvc2k}

\title{BrainBind: Bridging Synchronized EEG and fMRI through 
Long-Form Naturalistic Video Representations}

\addauthor{Shixiao Wang}{sxw1238@student.bham.ac.uk}{1}
\addauthor{Rining Wu}{ml20r2w@leeds.ac.uk}{1,2}
\addauthor{Jian K. Liu}{j.liu.22@bham.ac.uk}{1,2}

\addinstitution{
 School of Computer Science, \\ 
 Center for Human Brain Health, \\
 University of Birmingham, UK
}
\addinstitution{
 School of Computer Science, \\ 
 University of Leeds, UK
 }

\runninghead{Wang, Wu, and Liu}{BrainBind}

\usepackage{amsmath,amssymb,amsfonts}
\usepackage{booktabs}
\usepackage{array}
\usepackage{multirow}
\usepackage{graphicx}
\usepackage{xcolor}
\usepackage{url}
\usepackage{microtype}
\usepackage{placeins}
\hypersetup{pdfauthor={Shixiao Wang, Rining Wu, Jian K. Liu},
    pdftitle={BrainBind: Bridging Synchronized EEG and fMRI through Long-Form Naturalistic Video Representations},
    pdfborder={0 0 0}}

\newcommand{\eeg}{\mathrm{EEG}}
\newcommand{\fmri}{\mathrm{fMRI}}

\newcommand{\proxy}{\mathrm{S}}

\begin{document}
\maketitle
\begingroup
\renewcommand{\thefootnote}{}
\endgroup

\begin{abstract}
    Real-world vision is dynamic and scene-rich, while EEG and fMRI resolve it
    at complementary temporal and spatial scales.  Most neural decoding systems,
    however, align one modality in isolation and mostly against static images or
    short clips.  We study the synchronized case: EEG and fMRI are recorded
    during the same long-form naturalistic viewing, giving a shared stimulus but
    also recurring narrative context and many semantically similar candidates.
    We present \textbf{BrainBind}, a framework that maps synchronized EEG and
    fMRI into frozen long-form video embedding spaces and treats those spaces as
    common visual-semantic references, not literal neural codes.  On CineBrain,
    a five-seed Recall@K $2\!\times\!4$ matrix across two video anchors and four
    system configurations shows distinct retrieval profiles.  With
    Omni, fMRI organizes the full candidate list most strongly
    ($\mathrm{MRR}=23.82$, $\mathrm{nDCG@10}=25.31$), while Joint-Dual gives the
    highest mean label-aware R@1 ($13.52\%$) with a distinct rank-wide profile.  The
    complementary episode-disjoint evaluation quantifies the contribution of
    repeated stimulus identity and narrative context to the random split.  Across these
    experiments, proxy choice and system configuration produce distinct
    top-rank and rank-wide retrieval profiles within the present synchronized
    dataset.  The code is available at \href{https://github.com/Forainest789/Brainbind}{github.com/Forainest789/Brainbind}.
\end{abstract}

\section{Introduction}
\label{sec:intro}

Foundation visual representations have become an effective interface between brain activity and natural stimuli \cite{wen_neural_2017, zhang_reconstruction_2020, leonardis_aligning_2025, chen_decoding_2024}. Recent systems align fMRI or EEG with pretrained visual-semantic spaces before retrieval or reconstruction
\cite{dreamdiffusion,eegclip,brainvis,neuroclips,animateyourthoughts,mindeye,mindeye2,niceeeg}.
EEG has fast temporal resolution but weak spatial localization; fMRI localizes
cortical regions but integrates slow hemodynamic responses.  These
complementary measurements are nevertheless usually examined separately.  This
alignment strategy replaces a difficult pixel-level target with a structured
representation learned from large-scale visual data.  When EEG and fMRI are
recorded synchronously during the same stimulus, their alignment to a common
reference space tests complementary measurements while holding the dataset,
task, and stimulus set fixed.

Long-form naturalistic video makes this question both useful and difficult.
Nearby clips can share characters, locations, actions, and narrative context
\cite{hasson2010,hasson2015}; consequently, a random sample split may test
fine-grained discrimination within recurring context as much as generalization
to a new episode.  The candidate pool is also semantically dense, so top-rank
retrieval and rank-wide organization need not tell the same story.  We therefore
separate these views with Recall@K, MRR, mAP, and
nDCG, and pair the standard random split with correspondence controls and an
episode-disjoint evaluation.

\begin{figure*}[t]
    \centering
    \IfFileExists{images/overview_v2.pdf}{%
        \includegraphics[clip,width=1\textwidth]{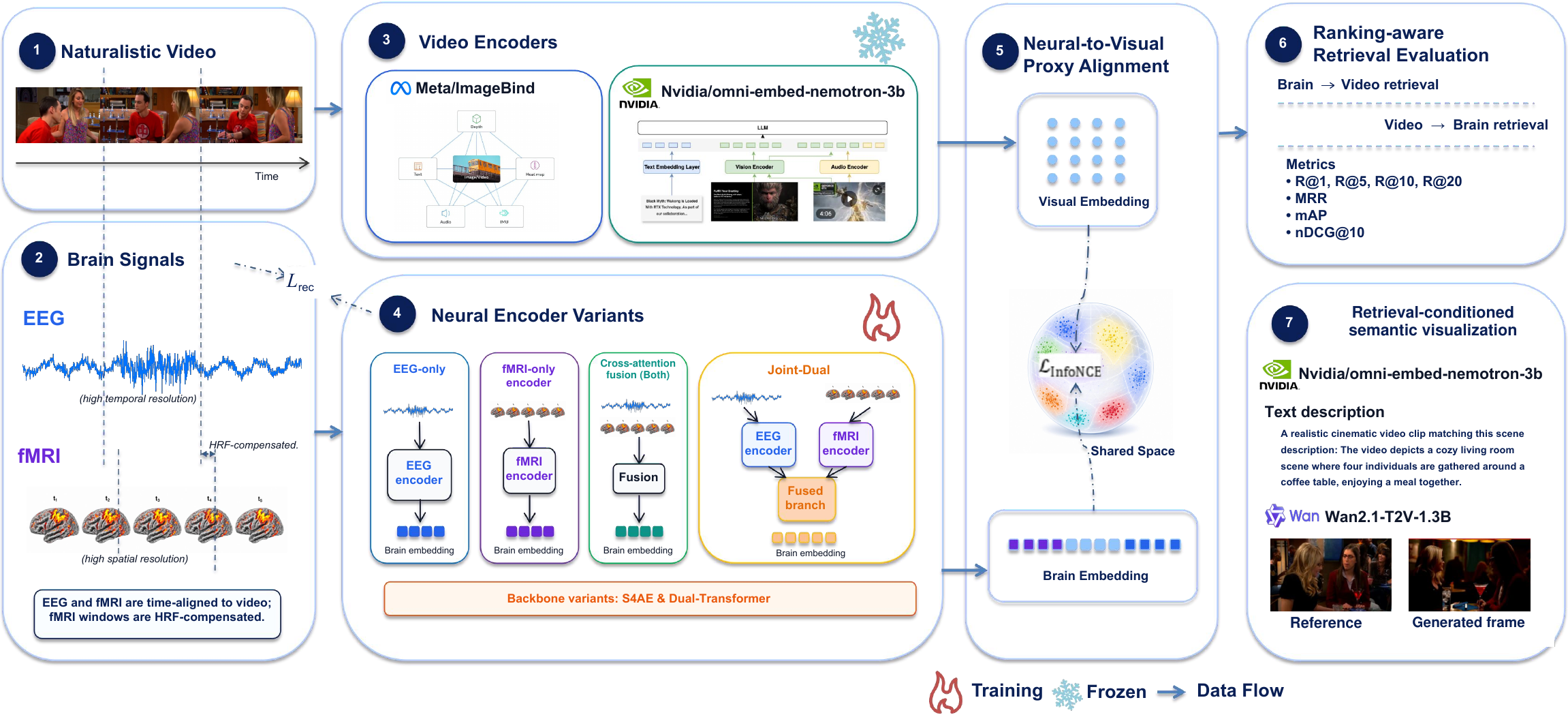}%
    }{%
        \fbox{\begin{minipage}[c][0.22\textheight][c]{0.96\textwidth}
                \centering
                \textbf{Figure placeholder: BrainBind overview.}
            \end{minipage}}%
    }
    \caption{Overview of BrainBind.  Synchronized EEG and fMRI responses are mapped
        into a frozen long-form video proxy space, which provides a shared
        visual-semantic geometry for comparing modality-specific and fused
        neural--video retrieval.  The dashed reconstruction path denotes the optional
        $L_1$ reconstruction loss $\mathcal{L}_{\mathrm{rec}}$ used only in designated
        diagnostic configurations; the core systems use contrastive-only training.
        The rightmost branch is retrieval-conditioned
        text-to-video visualization rather than direct neural reconstruction.}
    \label{fig:overview}
\end{figure*}

We introduce \textbf{BrainBind}, a representation-and-evaluation framework for
synchronized EEG and fMRI under long-form naturalistic viewing
(Figure~\ref{fig:overview}).  BrainBind maps each neural input to a frozen video
embedding space and treats that space as a shared visual-semantic reference
geometry.  The reference makes modality inputs, system configurations, and proxy
anchors comparable under the same stimulus and retrieval protocol; it is not
presented as a literal neural code.  We evaluate
BrainBind on CineBrain \cite{cinebrain}, using a five-seed Recall@K $2\!\times\!4$ matrix
 spanning two video anchors and four system configurations, together with
 geometry diagnostics, targeted S4AE analyses, correspondence controls, an
 episode-disjoint split, a three-seed loss/temperature ablation, and selective
 two-subject held-out-training checks.

On the random split, fMRI with the Omni anchor produces the strongest MRR, mAP, and
nDCG@10, indicating the strongest rank-wide organization.  The complete
Joint-Dual system instead produces the highest mean R@1 while remaining below fMRI on
those rank-wide metrics, revealing a precision--breadth trade-off.  EEG-only
alignment remains above control levels and exhibits a retrieval profile distinct from fMRI in the same reference space.  The
choice of proxy also matters: spectral and cross-anchor analyses associate the
broader Omni geometry with stronger neural alignment.  The sharp drop under
episode-disjoint evaluation then exposes the contribution of recurring
narrative context to random-split retrieval.

\paragraph{Contributions.}
BrainBind is a representation and evaluation framework for synchronized
EEG--fMRI alignment under long-form naturalistic video.  Its contributions are
threefold:

\noindent \textbf{Shared neural--video proxy formulation.}  We cast naturalistic
          neural decoding as stimulus-aligned neural-to-proxy representation
          learning, using frozen long-form video embeddings as a common
          visual-semantic geometry that organizes heterogeneous neural
          measurements without assuming literal proxy-to-neural correspondence.
          
\noindent \textbf{Proxy and system-configuration study.}  Across two video anchors,
          four system configurations, and five Recall@K seeds, we show that proxy
          choice and complete-system design yield distinct top-rank and rank-wide
          profiles: fMRI gives the strongest rank-wide organization, whereas
          Joint-Dual gives the highest mean label-aware R@1 on the random split.
          
\noindent \textbf{Ranking-aware evaluation and geometry diagnostics.}  We complement Recall@K with MRR, mAP, and nDCG@10, and augment retrieval
          evaluation with random-brain and shuffled controls, a three-seed
          loss and temperature ablation, selective held-out-training checks,
          episode disjoint evaluation, embedding-geometry diagnostics, and
          qualitative retrieval conditioned visualization, enabling joint
          analysis of top-rank retrieval and rank-wide organization.

\section{Related Work}
\label{sec:related}

\paragraph{fMRI decoding for static and dynamic visual stimuli.}
A large body of work aligns fMRI to pretrained visual or semantic embeddings as
a precursor to retrieval or generation.  For static images, MindEye
\cite{mindeye} and MindEye2 \cite{mindeye2} align fMRI to CLIP-like embeddings
and apply diffusion priors for reconstruction on large-scale fMRI datasets
\cite{nsd}; related systems explore semantic alignment
\cite{brainclip,takagi2023,umbrae} and low-data regimes \cite{mindeye2}.  For
dynamic stimuli, MinD-Video \cite{mindvideo}, NeuroClips \cite{neuroclips},
Brain Netflix \cite{brainnetflix}, Animate Your Thoughts
\cite{animateyourthoughts} extend this
paradigm to fMRI-driven video synthesis with diffusion-based decoders.  These
systems primarily study fMRI in isolation; BrainBind instead asks how
synchronized EEG and fMRI behave under the same long-form video geometry.

\paragraph{EEG decoding into visual representation spaces.}
EEG-based systems leverage visual and textual embedding spaces for image
generation, classification, and retrieval.  DreamDiffusion \cite{dreamdiffusion}
synthesizes images from EEG via diffusion priors; BrainVis \cite{brainvis} and
EEG-CLIP \cite{eegclip} bridge EEG to CLIP-like geometries.  NICE-EEG
\cite{niceeeg} establishes zero-shot EEG-to-image retrieval on THINGS-EEG
\cite{thingsdata}, and ATM \cite{atm} introduces adaptive masked transformers
for EEG-visual decoding.  Self-supervised EEG pretraining
\cite{eegformer,bendr,lbm} provides general-purpose temporal backbones.  These
studies are predominantly single-modality and often use static-image stimuli.
BrainBind addresses the controlled interaction of synchronized EEG, fMRI, and
frozen video proxies under long-form dynamic stimuli.

\paragraph{Multimodal foundation embeddings as stimulus geometries.}
Contrastive representation learning organizes heterogeneous modalities in shared
metric spaces \cite{cpc,moco,simclr,clip}, and ImageBind \cite{imagebind}
extends this principle to image, video, audio, text, and other sensory signals.
Retrieval-oriented video embeddings, including Omni-style models
\cite{omni,nvretriever} and temporally rich video encoders
\cite{internvideo,videoclip}, encode objects, actions, scene layout, and
temporal continuity.  For neural decoding, such embeddings serve not only as
supervision targets but also as reference geometries with different spectral
and neighborhood structure.  BrainBind explicitly compares proxy anchors and
treats their geometry as an experimental design variable, relating retrieval
differences to PCA spectral statistics and CCA analysis without equating either
proxy with a neural code.

\paragraph{Temporal modeling, contrastive learning, and ranking metrics.}
Transformers \cite{transformer} and structured state-space models \cite{s4,mamba}
provide complementary inductive biases for sequence modeling, and InfoNCE
\cite{cpc,simclr,moco} is the standard contrastive alignment objective, with
supervised and factorized extensions for richer positive structure
\cite{supcon,factorizedcl,positivepairs}.  BrainBind uses these established
components as controlled implementations.  Its targeted S4AE analyses examine
state-space configurations alongside Transformer encoders; they are not
isolated tests of temporal mixing because capacity and, in one diagnostic, the
reconstruction objective also change.
Beyond Recall@K, ranking-aware metrics such as MRR, mAP, and nDCG \cite{ndcg}
are essential for naturalistic retrieval where semantically proximate candidates
require analysis of the full ranked list in addition to top-1 accuracy.

\section{Method}
\label{sec:method}

\paragraph{Problem formulation.}
Each sample is a tuple
$(\mathbf{X}^{\eeg},\mathbf{X}^{\fmri},\mathbf{V})$, where
$\mathbf{X}^{\eeg}\in\mathbb{R}^{C\times T}$ is an EEG window,
$\mathbf{X}^{\fmri}$ is the corresponding fMRI observation after temporal
alignment for the hemodynamic response, and $\mathbf{V}$ is the matched video.
The goal is to learn modality-specific encoders that map EEG and fMRI into a
common video-grounded embedding space supporting bidirectional neural-to-video
and video-to-neural retrieval.

\paragraph{Long-form video proxy space.}
For anchor $a$ (Omni or ImageBind), a frozen video encoder $F_a$ produces a
pooled clip embedding.  We retain its native dimension and apply $\ell_2$
normalization:
\begin{equation}
    \mathbf{z}^{\proxy}=
    \mathrm{norm}_2\!\left(\mathrm{pool}(F_a(\mathbf{V}))\right).
    \label{eq:proxy}
\end{equation}
The neural output mapping matches the selected native target space
(2{,}048-D Omni or 1{,}024-D ImageBind); no video-side cross-anchor projection
is used.  Thus the native proxy geometry analyzed below is also the geometry
used for contrastive alignment.
We use $\mathbf{z}^{\proxy}$ solely as a structured stimulus reference.  The
experiment tests whether its geometric organization provides a useful inductive
bias for neural alignment; individual proxy dimensions carry no assumed neural
meaning.

\paragraph{Frozen reference design.}
We keep the video proxy encoder frozen throughout training.  This makes the
proxy geometry an experimental variable rather than a trainable component, and
it isolates changes in neural-to-proxy alignment from changes in the reference
space itself.  The stable reference delegates subject-specific adaptation to
the neural encoders and preserves comparability across modalities, fusion
designs, and anchors.

\paragraph{Neural encoders.}
Each modality $m\in\{\eeg,\fmri\}$ is encoded by a modality-specific encoder
$E_m$ followed by a projection head $g_m$, producing an $\ell_2$-normalized
embedding $\mathbf{z}^{m}$.  EEG-only, fMRI-only, and Both use Transformer
encoders \cite{transformer} with 512 hidden units, 8 heads, GELU, dropout 0.1,
and pre-norm.  In Both, each modality is encoded and mean-pooled before one
cross-attention module fuses the pair ahead of the final proxy-space mapping.  Joint-Dual uses
two independent S4AE encoders with mean fusion.  Each S4AE branch applies a
TemporalMLP and stacked blocks that pair a structured state-space time mixer
\cite{s4} with a channel-mixing MLP, both under residual connections and
layer normalization, before bottleneck projection.  A mirrored decoder
reconstructs the input when reconstruction is enabled.  For EEG these blocks
mix along the sampled time axis; in the fMRI configuration the five TRs are input channels and the structured mixer operates
along the voxel axis.  The primary matrix is therefore a controlled system
comparison across anchors and complete system configurations, not a
backbone-matched fusion ablation.

\paragraph{Proxy alignment and multimodal fusion.}
For a batch of $B$ matched pairs, define the directed InfoNCE term
\begin{equation}
    \ell(m,\proxy)
    =-\frac{1}{B}\sum_{i=1}^{B}
    \log
    \frac{\exp(\mathbf{z}^{m\top}_i\mathbf{z}^{\proxy}_i/\tau)}
    {\sum_{j=1}^{B}\exp(\mathbf{z}^{m\top}_i\mathbf{z}^{\proxy}_j/\tau)},
    \label{eq:nce}
\end{equation}
and optimize the symmetric loss
$\mathcal{L}_{\mathrm{NCE}}(m,\proxy)=
\tfrac12[\ell(m,\proxy)+\ell(\proxy,m)]$ with $\tau=0.05$.
The core matrix uses paired positives during training; the SupCon ablations use
all samples sharing the same \texttt{stim\_label} as positives.  Joint-Dual aligns the mean-fused embedding
and both modality-specific embeddings with the same video anchor:
\begin{equation}
    \mathcal{L}_{\mathrm{JD}}
    =\lambda_f\mathcal{L}_{\mathrm{NCE}}(\mathrm{fused},\proxy)
    +\lambda_m\!\left(\mathcal{L}_{\mathrm{NCE}}(\eeg,\proxy)
    +\mathcal{L}_{\mathrm{NCE}}(\fmri,\proxy)\right),
    \label{eq:jd}
\end{equation}
with $\lambda_f=1.0$ and $\lambda_m=0.5$, down-weighting the modality-specific
terms to prevent them from dominating the fused objective and preserving
complementary alignment paths for each modality.

\paragraph{S4AE diagnostics and reconstruction.}
S4AE denotes the backbone; reconstruction is enabled only for designated
diagnostic configurations.  The core
Joint-Dual model and its reduced-width retest are contrastive-only
($\alpha=0$).  In the targeted EEG--ImageBind S4AE diagnostic, reconstruction
is enabled for the active modality:
\begin{equation}
    \mathcal{L}_{\mathrm{S4AE}}=
    \beta\mathcal{L}_{\mathrm{NCE}}+
    \alpha\mathcal{L}_{\mathrm{rec}}, \quad
    \mathcal{L}_{\mathrm{rec}}=
    \frac{1}{|\mathbf{X}^{m}|}\|\hat{\mathbf{X}}^{m}-\mathbf{X}^{m}\|_1.
    \label{eq:s4ae}
\end{equation}
All reported configurations set $\beta=1$.  The EEG--ImageBind diagnostic sets $\alpha=0.25$; the core and reduced-width
Joint-Dual systems set $\alpha=0$.
The reported S4AE rows provide targeted configuration evidence across width,
modality, and objective choices.

Given an EEG or fMRI window, the S4AE branch first lifts the input with a
lightweight TemporalMLP, then applies stacked blocks that combine a structured state-space
time mixer with a channel-mixing MLP.  For EEG the structured mixer operates
along sampled time; for fMRI the five TRs are input channels and the 18{,}926
voxels form the mixed sequence axis, so we describe the fMRI operation as
structured state-space mixing rather than temporal mixing.  When reconstruction
is enabled, mirrored decoder blocks map the latent sequence back to the active
neural input.

\paragraph{Evaluation protocol.}
All candidates are ranked by cosine similarity.  We compute Recall@K
($K\in\{1,5,10,20\}$), MRR, mean average precision (mAP), and nDCG@10
\cite{ndcg} from the same ranked
list.  All candidates with the same \texttt{stim\_label} are positives.  This label-aware protocol
reflects the repeated-observation structure of naturalistic EEG--fMRI data and
is applied identically to every learned configuration and label-aware control.
The one-to-one shuffled-gallery control is defined separately below.  We also
report effective rank, $k$-nearest-neighbor density, and a densest-1\%
neighborhood audit to characterize the learned neural embedding geometry.

\section{Experiments}
\label{sec:experiments}

\paragraph{Dataset, protocol, and experimental matrix.}
We evaluate on CineBrain \cite{cinebrain}, a publicly available synchronized
EEG--fMRI dataset collected during long-form naturalistic audiovisual viewing
with HRF-compensated fMRI alignment.  A fixed 60/20/20 random sample split is
shared across all configurations, and retrieval is evaluated over the full
evaluation split.  With approximately four observations per stimulus, the
expected probability that a test sample shares its exact segment with training
is $1-0.4^3=93.6\%$.  We use this random split as an in-distribution retrieval
setting and complement it with episode-disjoint evaluation.  All eight core
configurations have five-seed Recall@K results: two frozen proxy anchors, Omni \cite{omni} and ImageBind
\cite{imagebind}, crossed with EEG-only, fMRI-only, cross-attention fusion
(Both), and Joint-Dual.  The first three systems use Transformer encoders;
Joint-Dual uses two S4AE branches and mean fusion.  In the label-aware setting,
all same-\texttt{stim\_label}
candidates are treated as positives, and R@K, MRR, mAP, and nDCG@10 are
 computed from the same ranked list.  EEG--ImageBind MRR, mAP, and nDCG@10 use
 four available seeds because one run lacks ranking outputs; all other core
 ranking aggregates use five.  We additionally report random-brain and
 shuffled-video controls, targeted S4AE configuration diagnostics, a three-seed
 episode-disjoint loss and temperature ablation, selective two-subject
  held-out-training checks, and an episode-disjoint stress test.

Each sample pairs a video segment with its corresponding EEG and fMRI windows
under the dataset-provided temporal alignment, including hemodynamic-response
compensation for fMRI.  This common temporal reference keeps stimulus alignment
fixed across modality, anchor, fusion, and backbone comparisons.  BrainBind treats frozen visual
representations as proxy stimulus geometries: structured anchors for neural
alignment, not literal neural codes.

For each retrieval direction, brain-to-video uses a neural embedding as the query
against the video proxy gallery, while video-to-brain reverses this direction.
The theoretical chance and random-brain rows use label-aware repeated positives.
The shuffled-video row instead evaluates
one-to-one row correspondence after independently permuting the video gallery.
All controls retain the same gallery size and ranking metrics.  The
episode-disjoint split assigns each season--episode group to one partition,
with 19{,}444/6{,}480/6{,}480 as train, validation, and test pairs.

\paragraph{Implementation details.}
Neural outputs match the selected anchor's native dimension (2{,}048 for Omni
and 1{,}024 for ImageBind) and are $\ell_2$-normalized.
Transformer branches use 512 hidden units, 8 heads, GELU, dropout 0.1, and
pre-norm; fMRI inputs are 5~TR observations projected from 18,926 voxels; EEG
uses temporal stride~8.  Both applies one cross-attention module after
modality-specific encoding and mean pooling.  Joint-Dual uses independent S4AE
branches, mean fusion, and contrastive-only training.  All runs use AdamW \cite{adamw} with
cosine annealing \cite{sgdr}, mixed precision, gradient clipping, and
contrastive training.  Checkpoints are selected by validation mean
R@1.  Statistical comparisons use two-sided paired $t$-tests over five matched
random-split seeds; $p$-values are unadjusted.  Three-seed diagnostics are
interpreted descriptively.  Additional architecture,
control, and compute details are in the supplementary.

\subsection{Training and compute}
The primary Transformer systems contain 47.8--49.9M parameters;
Joint-Dual--Omni has 293.8M.  All runs use two NVIDIA A100 80GB GPUs,
PyTorch~2.5.1 with CUDA~12.1, and the Omni Transformer runs use bf16 mixed
precision.  Table~\ref{tab:main_training} reports measured training duration and
validation recall for the runs with recorded timing.

\begin{table}[!htbp]
    \centering
    \scriptsize
    \resizebox{\columnwidth}{!}{%
        \begin{tabular}{llr}
            \toprule
            Experiment family             & Backbone                         & Total parameters \\
            \midrule
            EEG / fMRI / Both, Omni       & Transformer~\cite{transformer}   & 49,907,712       \\
            EEG / fMRI / Both, ImageBind  & Transformer~\cite{transformer}   & 47,809,536       \\
            Joint-Dual, Omni              & 2$\times$S4AE (structured SSM)~\cite{s4} & 293,835,134      \\
            Joint-Dual, ImageBind         & 2$\times$S4AE (structured SSM)~\cite{s4} & 191,595,902      \\
            EEG S4AE-matched, Omni        & S4AE (structured SSM)~\cite{s4} & 118,541,167      \\
            fMRI S4AE-matched, Omni       & S4AE (structured SSM)~\cite{s4} & 118,541,167      \\
            Joint-Dual S4AE-matched, Omni & S4AE (structured SSM)~\cite{s4} & 142,754,222      \\
            \bottomrule
        \end{tabular}%
    }
    \caption{Model parameter counts for the reported configurations.}
    \label{tab:main_params}
\end{table}

Joint-Dual--Omni is nearly six times larger than the fMRI-only Omni Transformer.
This capacity difference positions the result as a system-level comparison:
Joint-Dual more often places a label-positive candidate first, while fMRI-only
organizes broader candidate neighborhoods more strongly.  A capacity-matched
control is a natural extension of this analysis.

\begin{table*}[t]
    \centering
    \scriptsize
    \resizebox{\textwidth}{!}{%
        \begin{tabular}{lccccc}
            \toprule
            Experiment                        & Runs & Avg.\ epochs & Speed (it/s) & Duration (min/run) & Best val mean R@1 (\%) \\
            \midrule
            EEG--Omni Transformer             & 5    & 50.0        & 0.067        & 103.50             & 5.17                   \\
            fMRI--Omni Transformer            & 5    & 50.0        & 0.082        & 83.14              & 22.33                  \\
            Both--Omni Transformer            & 5    & 50.0        & 0.063        & 111.30             & 20.69                  \\
            Joint-Dual--Omni 2$\times$S4AE    & 5    & 50.0        & 0.012        & 585.22             & 16.54                  \\
            EEG--Omni S4AE-matched            & 5    & 24.4        & 0.041        & 82.44              & 1.10                   \\
            fMRI--Omni S4AE-matched           & 5    & 46.8        & 0.027        & 357.98             & 16.76                  \\
            Joint-Dual--ImageBind 2$\times$S4AE & 5  & 23.0        & 0.027        & 117.30             & 11.12                  \\
            Reduced J-Dual--Omni S4AE         & 2    & 50.0        & 0.014        & 508.30             & 18.87                  \\
            \bottomrule
        \end{tabular}%
    }
    \caption{Training duration and throughput for runs with recorded timing.}
    \label{tab:main_training}
\end{table*}

\section{Results}
\label{sec:results}

\subsection{Random-split retrieval reveals asymmetric modality strength}
Tables~\ref{tab:main_recall_matrix} and~\ref{tab:main_ranking_matrix} summarize
the label-aware bidirectional retrieval matrix.  With Omni, fMRI reaches mean
R@1 12.71, MRR 23.82, mAP 21.37, and nDCG@10 25.31, making it the strongest
standalone modality across the ranked list.  EEG aligned to the same proxy
reaches mean R@1 1.91 in a 6{,}480 neural--video sample-pair pool where candidates share characters,
scenes, and narrative context.  Its absolute retrieval is modest, yet its
separation from random-brain and shuffled-video controls indicates measurable
neural--video correspondence on the random split.  The episode-disjoint
experiment below tests how much remains when recurring narrative context no
longer crosses partitions.

\begin{table}[!htbp]
    \centering
    \small
    \setlength{\tabcolsep}{6pt}
    \renewcommand{\arraystretch}{1.12}
\textbf{(a) Brain-to-video retrieval (s2v)}\par\smallskip
\begin{tabular}{llcccc}
\toprule
Anchor & Input & R@1 & R@5 & R@10 & R@20 \\
\midrule
Omni & EEG & 1.51{\small$\pm$0.13} & 6.09{\small$\pm$0.90} & 11.26{\small$\pm$1.55} & 20.16{\small$\pm$1.82} \\
Omni & fMRI & 9.81{\small$\pm$0.16} & 27.70{\small$\pm$0.64} & 42.71{\small$\pm$0.31} & 59.85{\small$\pm$0.53} \\
Omni & Both & 8.41{\small$\pm$0.30} & 25.08{\small$\pm$0.40} & 39.34{\small$\pm$0.41} & 56.04{\small$\pm$0.69} \\
Omni & J-Dual & \textbf{10.28}{\small$\pm$0.24} & 17.47{\small$\pm$0.54} & 24.20{\small$\pm$1.17} & 33.35{\small$\pm$1.78} \\
\midrule
ImageBind & EEG & 0.65{\small$\pm$0.05} & 2.61{\small$\pm$0.11} & 4.77{\small$\pm$0.13} & 9.23{\small$\pm$0.26} \\
ImageBind & fMRI & \textbf{3.90}{\small$\pm$0.42} & 12.90{\small$\pm$0.82} & 20.97{\small$\pm$1.03} & 32.68{\small$\pm$0.86} \\
ImageBind & Both & 3.29{\small$\pm$0.20} & 10.49{\small$\pm$0.59} & 17.94{\small$\pm$0.90} & 28.96{\small$\pm$1.18} \\
ImageBind & J-Dual & 0.99{\small$\pm$0.12} & 3.52{\small$\pm$0.24} & 6.76{\small$\pm$0.50} & 11.99{\small$\pm$0.85} \\
\bottomrule
\end{tabular}
\par\medskip
\textbf{(b) Video-to-brain retrieval (v2s)}\par\smallskip
\begin{tabular}{llcccc}
\toprule
Anchor & Input & R@1 & R@5 & R@10 & R@20 \\
\midrule
Omni & EEG & 2.32{\small$\pm$0.45} & 10.16{\small$\pm$1.39} & 18.38{\small$\pm$1.87} & 30.65{\small$\pm$2.54} \\
Omni & fMRI & 15.61{\small$\pm$0.45} & 40.08{\small$\pm$0.62} & 56.29{\small$\pm$0.52} & 72.20{\small$\pm$0.57} \\
Omni & Both & 14.05{\small$\pm$0.62} & 38.04{\small$\pm$0.27} & 53.47{\small$\pm$0.60} & 69.16{\small$\pm$0.64} \\
Omni & J-Dual & \textbf{16.75}{\small$\pm$0.26} & 26.19{\small$\pm$0.70} & 33.46{\small$\pm$1.16} & 43.81{\small$\pm$1.64} \\
\midrule
ImageBind & EEG & 0.56{\small$\pm$0.06} & 3.07{\small$\pm$0.22} & 5.97{\small$\pm$0.55} & 11.31{\small$\pm$0.43} \\
ImageBind & fMRI & \textbf{8.35}{\small$\pm$0.66} & 24.05{\small$\pm$1.38} & 34.90{\small$\pm$1.24} & 48.18{\small$\pm$0.95} \\
ImageBind & Both & 6.23{\small$\pm$0.76} & 20.30{\small$\pm$1.63} & 30.93{\small$\pm$1.75} & 44.32{\small$\pm$1.55} \\
ImageBind & J-Dual & 1.87{\small$\pm$0.35} & 6.90{\small$\pm$0.63} & 11.65{\small$\pm$0.67} & 18.89{\small$\pm$0.83} \\
\bottomrule
\end{tabular}
    \caption{Label-aware CineBrain bidirectional retrieval matrix. Values are
        percentages, reported as mean$\pm$std across five seeds. Both directions
        use the same evaluation pair pool with query and gallery roles reversed.
        EEG and fMRI use single-modality Transformers; Both uses cross-attention
        fusion with a Dual-Transformer; J-Dual uses mean fusion with two S4AE
        branches. Best R@1 values within each anchor and direction are bold.}
    \label{tab:main_recall_matrix}
\end{table}

\begin{table}[!htbp]
    \centering
    \scriptsize
    \resizebox{\columnwidth}{!}{%
        \begin{tabular}{llllcccc}
            \toprule
            Anchor    & Input  & Fusion     & Backbone         & R@1                             & MRR                                  & mAP                                  & nDCG@10                              \\
            \midrule
            Omni      & EEG    & single     & Transformer      & 1.91{\scriptsize$\pm$0.28}           & 6.60{\scriptsize$\pm$0.58}           & 6.03{\scriptsize$\pm$0.49}           & 6.06{\scriptsize$\pm$0.75}           \\
            Omni      & fMRI   & single     & Transformer      & 12.71{\scriptsize$\pm$0.24}          & \textbf{23.82}{\scriptsize$\pm$0.29} & \textbf{21.37}{\scriptsize$\pm$0.19} & \textbf{25.31}{\scriptsize$\pm$0.20} \\
            Omni      & Both   & cross-attn & Dual-Transformer & 11.23{\scriptsize$\pm$0.30}          & 21.91{\scriptsize$\pm$0.18}          & 19.61{\scriptsize$\pm$0.11}          & 23.22{\scriptsize$\pm$0.20}          \\
            Omni      & J-Dual & mean       & 2$\times$S4AE    & \textbf{13.52}{\scriptsize$\pm$0.11} & 18.69{\scriptsize$\pm$0.35}          & 15.09{\scriptsize$\pm$0.39}          & 16.52{\scriptsize$\pm$0.51}          \\
            \midrule
            ImageBind & EEG    & single     & Transformer      & 0.61{\scriptsize$\pm$0.05}           & 2.83{\scriptsize$\pm$0.06}$^\dagger$ & 2.71{\scriptsize$\pm$0.06}$^\dagger$ & 2.20{\scriptsize$\pm$0.06}$^\dagger$ \\
            ImageBind & fMRI   & single     & Transformer      & \textbf{6.12}{\scriptsize$\pm$0.52}  & \textbf{13.27}{\scriptsize$\pm$0.67} & \textbf{11.41}{\scriptsize$\pm$0.53} & \textbf{12.99}{\scriptsize$\pm$0.72} \\
            ImageBind & Both   & cross-attn & Dual-Transformer & 4.76{\scriptsize$\pm$0.47}           & 11.23{\scriptsize$\pm$0.68}          & 9.78{\scriptsize$\pm$0.53}           & 10.94{\scriptsize$\pm$0.70}          \\
            ImageBind & J-Dual & mean       & 2$\times$S4AE    & 1.43{\scriptsize$\pm$0.22}           & 4.42{\scriptsize$\pm$0.31}           & 3.86{\scriptsize$\pm$0.21}           & 3.74{\scriptsize$\pm$0.28}           \\
            \bottomrule
        \end{tabular}
    }
    \caption{Ranking-aware evaluation for the controlled retrieval matrix.  Mean
        R@1, MRR, mAP, and nDCG@10 are averaged across the s2v and v2s
        directions.  $^\dagger$EEG--ImageBind ranking
        metrics are averaged over four seeds because one seed lacks ranking-metric
        outputs.}
    \label{tab:main_ranking_matrix}
\end{table}

\subsection{J-Dual changes the retrieval profile across ranks}
Joint-Dual--Omni achieves the highest mean R@1 (13.52) in the main matrix,
exceeding fMRI-only by $+0.81$\,pp ($p=8\times10^{-4}$).  fMRI--Omni,
however, provides stronger rank-wide
organization: it leads on MRR, mAP, and nDCG@10 by 5.13, 6.28, and 8.79\,pp
respectively ($p=8.5\times10^{-7}$, $9.7\times10^{-7}$, and
$1.2\times10^{-6}$).  Joint-Dual therefore places a label-positive candidate
first more often, whereas fMRI-only organizes positives more strongly across
the ranked list.  We describe this system-level pattern as a precision--breadth
trade-off.  Joint-Dual uses two
S4AE branches instead of the fMRI-only Transformer, is about six times larger
(294M vs.\ 50M parameters), and optimizes fused and modality-specific
contrastive terms; the reported gain therefore characterizes the complete
Joint-Dual system.  Architecture-, capacity-, and objective-matched controls
provide a focused direction for isolating individual contributions.
Full matched-seed comparisons and signed effect differences are reported in
Table~\ref{tab:main_stats}.

\begin{table}[!htbp]
    \centering
    \scriptsize
    \resizebox{\columnwidth}{!}{%
        \begin{tabular}{lccc l}
            \toprule
            Comparison                   & Metric   & $\Delta$ pp & $p$                  & Interpretation               \\
            \midrule
            Both vs.\ fMRI (Omni)        & mean R@1 & -1.48       & 0.0012               & System-level difference      \\
            Both vs.\ fMRI (Omni)        & nDCG@10  & -2.09       & $3.3{\times}10^{-5}$ & Rank-structure difference    \\
            J-Dual vs.\ fMRI (Omni)      & mean R@1 & +0.81       & 0.0008               & Top-rank difference          \\
            J-Dual vs.\ fMRI (Omni)      & MRR      & -5.13       & $8.5{\times}10^{-7}$ & Rank-organization difference \\
            J-Dual vs.\ fMRI (Omni)      & mAP      & -6.28       & $9.7{\times}10^{-7}$ & Candidate-space difference   \\
            J-Dual vs.\ fMRI (Omni)      & nDCG@10  & -8.79       & $1.2{\times}10^{-6}$ & Rank-structure difference    \\
            fMRI ImageBind vs.\ fMRI Omni & nDCG@10 & -12.32      & $1.3{\times}10^{-6}$ & Anchor-associated difference \\
            \bottomrule
        \end{tabular}%
    }
    \caption{Paired seed-level comparisons.  Positive $\Delta$ indicates the first
        configuration is higher.  The Joint-Dual rows quantify the difference
        between precise matching and rank-wide organization; the anchor row
        quantifies the consistent Omni advantage.}
    \label{tab:main_stats}
\end{table}

\begin{figure*}[t]
    \centering
    \setlength{\tabcolsep}{1.5pt}
    \renewcommand{\arraystretch}{0.82}
    \scriptsize
    \begin{tabular}{ccccc}
                                                                                                                              & fMRI--Omni & Both--Omni & J-Dual--Omni & EEG--Omni \\
        t-SNE                                                                                                                 &
        \includegraphics[clip,width=0.225\textwidth]{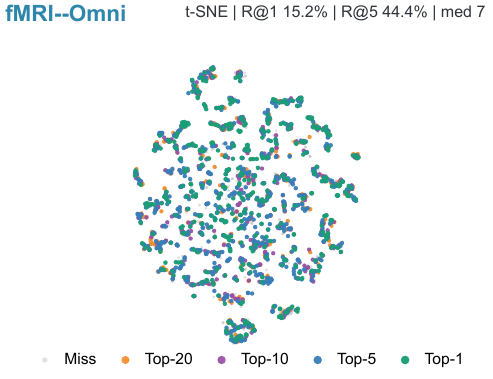}  &
        \includegraphics[clip,width=0.225\textwidth]{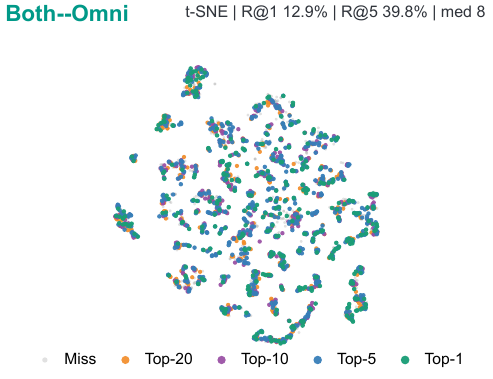}  &
        \includegraphics[clip,width=0.225\textwidth]{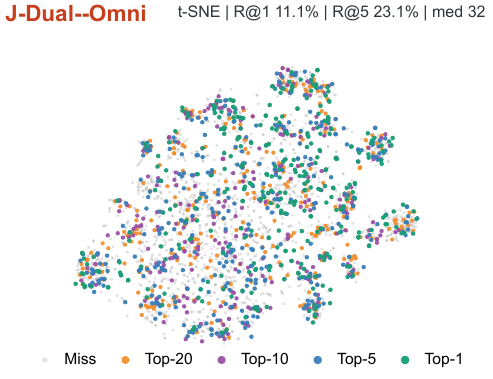} &
        \includegraphics[clip,width=0.225\textwidth]{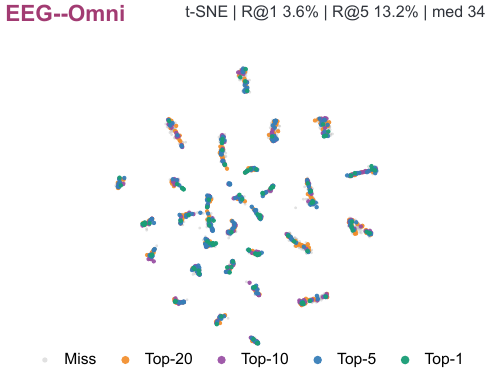}                                                        \\
        UMAP                                                                                                                  &
        \includegraphics[clip,width=0.225\textwidth]{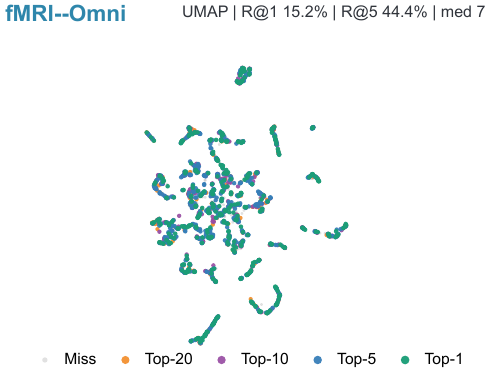}  &
        \includegraphics[clip,width=0.225\textwidth]{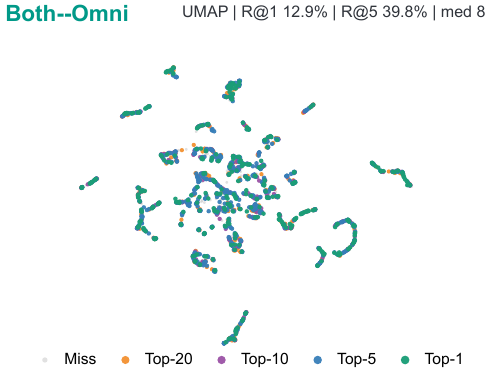}  &
        \includegraphics[clip,width=0.225\textwidth]{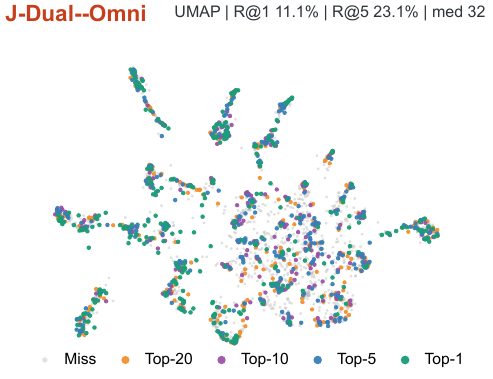} &
        \includegraphics[clip,width=0.225\textwidth]{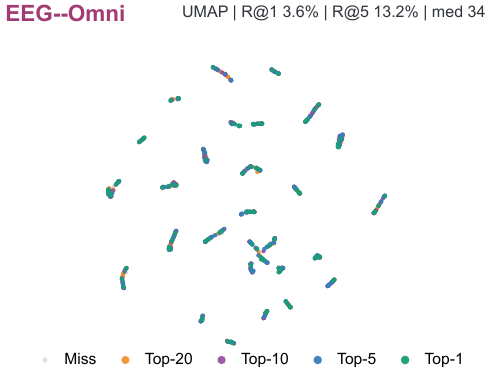}
    \end{tabular}

    \vspace{0.8mm}
    \begin{tabular}{ccccc}
        \includegraphics[clip,width=0.29\textwidth]{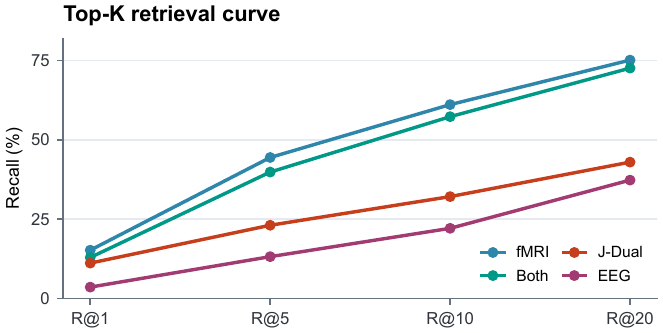} &
        \includegraphics[clip,width=0.16\textwidth]{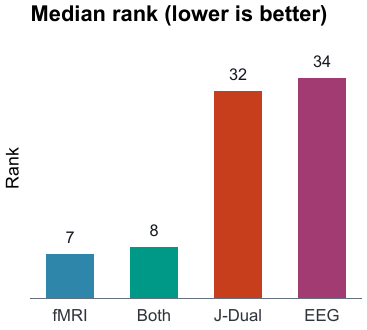}          &
        \includegraphics[clip,width=0.16\textwidth]{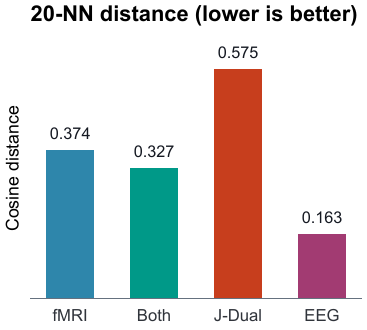}        &
        \includegraphics[clip,width=0.16\textwidth]{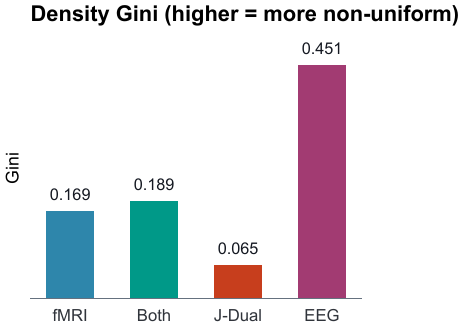}         &
        \includegraphics[clip,width=0.16\textwidth]{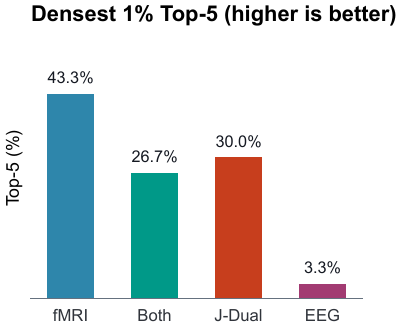}
    \end{tabular}

    \vspace{2mm}
    \caption{\textbf{Embedding-geometry diagnostics for Omni-aligned brain
            representations.}  The top block compares retrieval-colored 2D projections
        (t-SNE and UMAP) for the four Omni-aligned brain embeddings on a
        3{,}000-sample diagnostic subset. These subset metrics are distinct from
        the full-pool, multi-seed aggregates in Tables~\ref{tab:main_recall_matrix}
        and~\ref{tab:main_ranking_matrix}. The bottom
        block reports rank-space and high-dimensional neighborhood statistics:
        Top-K retrieval curve, median first-positive rank, 20-NN cosine distance,
        density Gini coefficient, and Top-5 retrieval within the densest 1\%
        neighborhoods.  Projections are qualitative; metric panels are computed in
        retrieval rank space or the original high-dimensional embedding space.
        Higher Gini values indicate stronger local density non-uniformity rather
        than a quality ranking.}
    \label{fig:ranking_geometry_tsne_umap}
\end{figure*}

The embedding diagnostics in Figure~\ref{fig:ranking_geometry_tsne_umap}
connect the 2D projections to high-dimensional neighborhood statistics.
fMRI--Omni exhibits the strongest rank-wide profile: the steepest Top-$K$ curve on
the 3{,}000-sample diagnostic subset (R@5 44.43, R@10 61.10) and the lowest
median first-positive rank (7).  Both--Omni remains close to fMRI--Omni but
slightly weakens rank-wide organization.  The R@1 ordering in this diagnostic
subset differs from the full evaluation matrix: fMRI--Omni reaches 15.2\%,
Both--Omni 12.9\%, J-Dual--Omni 11.1\%, and EEG--Omni 3.6\%.
J-Dual--Omni has a median first-positive rank of 32 on this subset.
Its highest mean R@1 in Table~\ref{tab:main_ranking_matrix} refers to the full
evaluation pool averaged over seeds, not to the subset shown in
Figure~\ref{fig:ranking_geometry_tsne_umap}.  EEG--Omni exhibits stronger local
density non-uniformity.  These diagnostics characterize the displayed subset;
the full-pool aggregates support the primary system comparisons.

\subsection{Proxy-anchor geometry is associated with alignment quality}
\begin{figure}[t]
    \centering
    \begin{tabular}{@{}c@{\hspace{0.02\columnwidth}}c@{}}
        \includegraphics[clip,width=0.49\columnwidth]{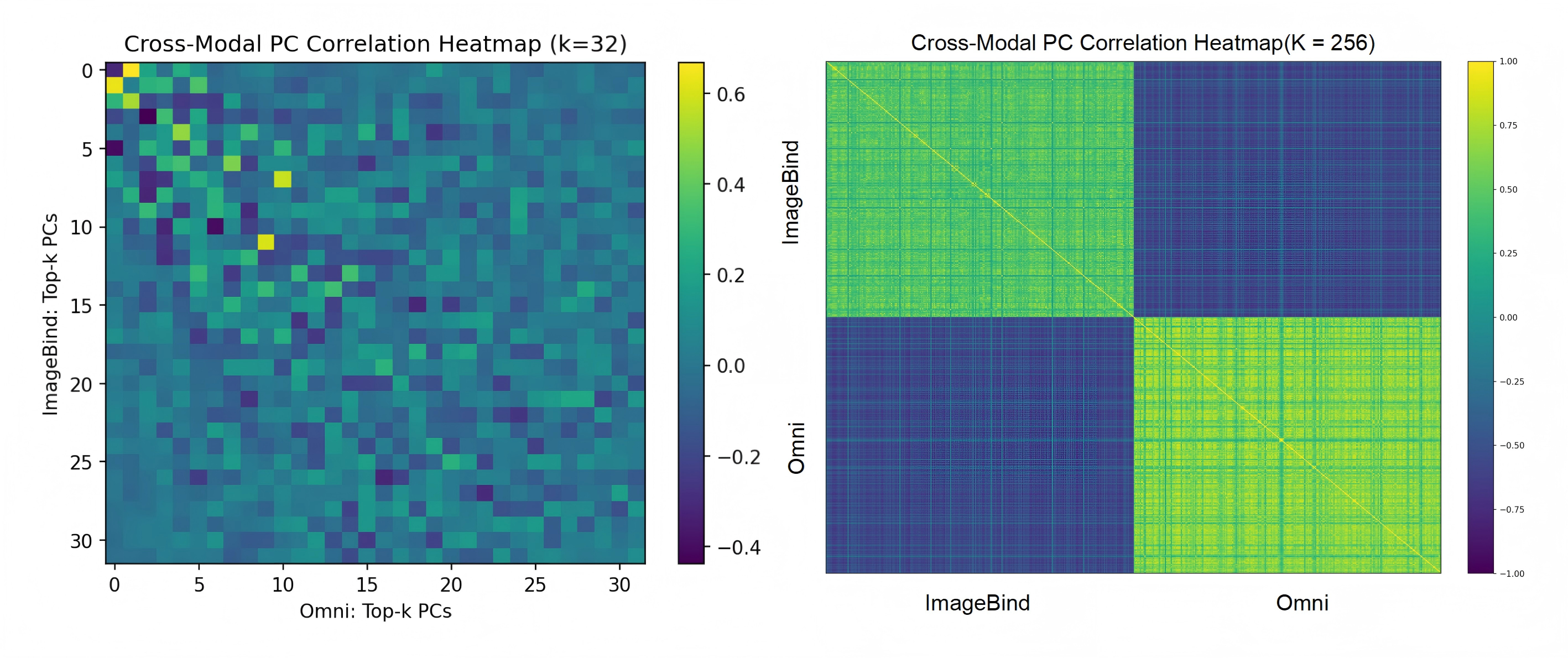} &
        \includegraphics[clip,width=0.49\columnwidth]{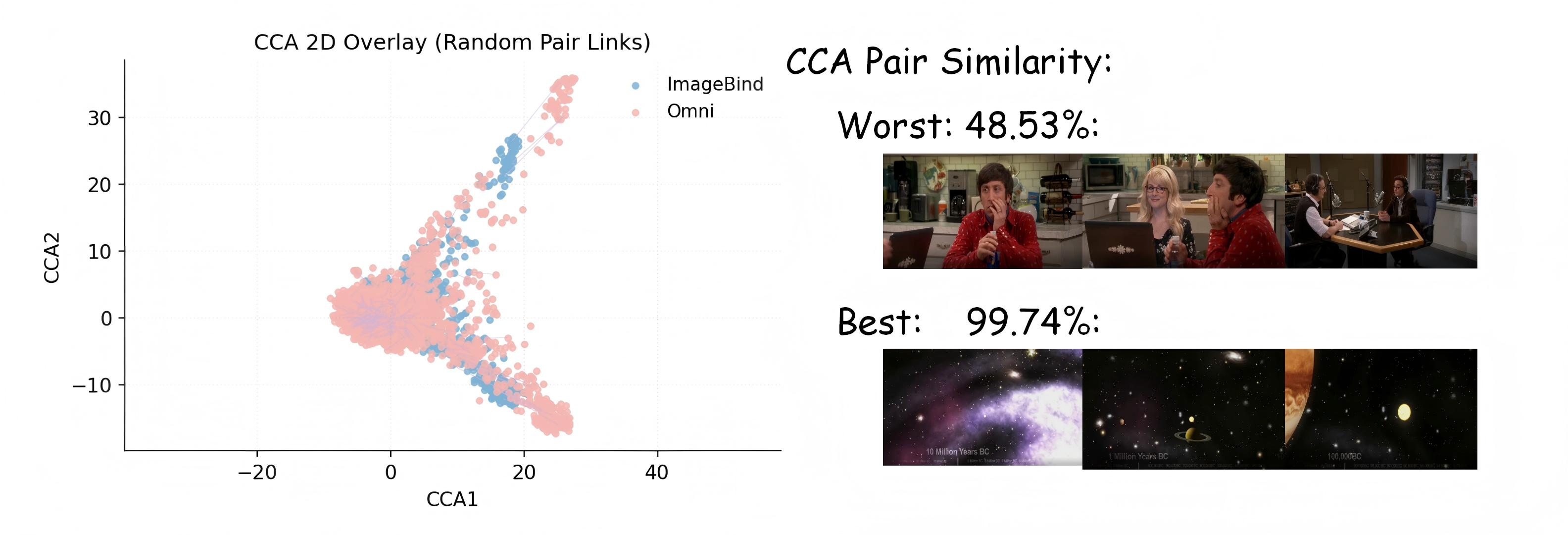}
    \end{tabular}
    \caption{\textbf{Cross-anchor geometry of the two frozen video proxies.}
        ImageBind concentrates more variance in fewer principal components;
        Omni has broader spectral support and lower redundancy.  Independently
        fitted PCA followed by in-sample CCA characterizes their linear correspondence.}
    \label{fig:cross_anchor_geometry}
\end{figure}

Omni outperforms ImageBind across all configurations.  For fMRI, Omni achieves
s2v R@1 of 9.81 versus 3.90 for ImageBind; the nDCG@10 gap is $12.32$\,pp
($p=1.3\times10^{-6}$).  Figure~\ref{fig:cross_anchor_geometry} and the
supplementary cross-anchor analysis characterize the corresponding geometric
difference.  Native-space spectral analysis shows that ImageBind concentrates 94.9\% of its variance in the top 256 PCA
components, whereas Omni distributes its variance more broadly across components
(86.9\% in the top 256), yielding a higher effective rank (1{,}112 vs.\ 426) and
lower feature redundancy.  This broader spectrum is consistent with the stronger
Omni retrieval results.  Exploratory CCA, applied after independently fitted
256-dimensional PCA representations and evaluated on the same 8{,}100 slices,
has mean correlation $0.88$ across the first 64 components.  This in-sample
correspondence motivates future multi-anchor analysis.  The consistent Omni advantage across all eight
systems supports a systematic relationship between proxy-anchor geometry and
alignment quality for both EEG and fMRI.

\subsection{Temporal and S4AE diagnostics characterize modeling behavior}
A temporal-order check reverses, shifts, and TR-shuffles the EEG sequence at
test time.  Each perturbation changes retrieval by at most 0.5 percentage
points, showing that the current mean-pooled EEG Transformer emphasizes
order-robust neural--video correspondence.  This result motivates sequence-aware
extensions for explicitly modeling temporal order.

The targeted S4AE rows add two configuration checks.  The core Joint-Dual model
already uses two S4AE branches and contrastive-only training.  A reduced-width
contrastive-only retest reaches mean R@1 14.38 over three seeds, compared with
13.52 for the larger five-seed core system.  This configuration-dependent pattern
motivates matched width analysis.  An EEG--ImageBind S4AE
configuration with reconstruction reaches mean R@1 0.15, below its Transformer
counterpart at 0.61.  Together, these rows identify promising S4AE
configurations and motivate a fuller modality-by-anchor study.

\subsection{Loss and temperature ablations under the episode-disjoint split}
We additionally complete two three-seed ablations under the episode-disjoint
protocol.  The first varies the EEG-only S4AE
objective: InfoNCE versus SupCon, crossed with contrastive-only versus
reconstruction-assisted training.  SupCon defines positives by shared stimulus
labels across neural and video views.  In the contrastive-only setting, SupCon
improves the EEG baseline across recall and ranking metrics, including mean
R@10 $0.782\%$ versus $0.532\%$ and mean R@20 $1.258\%$ versus $0.977\%$.
The reconstruction-assisted SupCon variant reaches the highest mean R@1
($0.136\pm0.072\%$) but has larger seed variation and is not consistently
stronger at R@5, R@10, or R@20 than the contrastive-only SupCon row.

The second ablation uses a separate EEG Transformer configuration, holds the
SupCon reconstruction-assisted objective fixed, reconstructs the mean-pooled
512-dimensional projected EEG input, and varies the contrastive
temperature on the episode split.  The configured default
$\tau{=}0.05$ is the strongest three-seed setting on every reported metric:
mean R@1 $0.180\%$, R@5 $0.486\%$, R@10 $0.936\%$, R@20 $1.674\%$, MRR
$0.614\%$, and mAP $0.486\%$.  Both ablations use three coupled model/split-seed
runs, providing complementary configuration evidence across episode assignments.

Tables~\ref{tab:main_loss_ablation} and~\ref{tab:main_temperature} report the
completed loss and temperature sweeps.

\begin{table}[!htbp]
    \centering
    \scriptsize
    \setlength{\tabcolsep}{3.0pt}
    \renewcommand{\arraystretch}{1.08}
    \resizebox{\columnwidth}{!}{%
        \begin{tabular}{lcccccc}
            \toprule
            Objective & R@1 & R@5 & R@10 & R@20 & MRR & mAP \\
            \midrule
            InfoNCE, contrastive-only        & 0.080{\scriptsize$\pm$0.025} & 0.345{\scriptsize$\pm$0.047} & 0.532{\scriptsize$\pm$0.078} & 0.977{\scriptsize$\pm$0.105} & 0.393{\scriptsize$\pm$0.028} & 0.343{\scriptsize$\pm$0.009} \\
            SupCon, contrastive-only         & 0.111{\scriptsize$\pm$0.038} & 0.448{\scriptsize$\pm$0.008} & 0.782{\scriptsize$\pm$0.047} & 1.258{\scriptsize$\pm$0.088} & 0.477{\scriptsize$\pm$0.023} & 0.376{\scriptsize$\pm$0.030} \\
            InfoNCE, reconstruction-assisted & 0.080{\scriptsize$\pm$0.025} & 0.316{\scriptsize$\pm$0.074} & 0.689{\scriptsize$\pm$0.103} & 1.175{\scriptsize$\pm$0.171} & 0.435{\scriptsize$\pm$0.057} & 0.391{\scriptsize$\pm$0.020} \\
            SupCon, reconstruction-assisted  & \textbf{0.136}{\scriptsize$\pm$0.072} & 0.399{\scriptsize$\pm$0.098} & 0.648{\scriptsize$\pm$0.167} & 1.052{\scriptsize$\pm$0.161} & 0.464{\scriptsize$\pm$0.081} & 0.375{\scriptsize$\pm$0.028} \\
            \bottomrule
        \end{tabular}%
    }
    \caption{EEG-only S4AE loss ablation on the episode-disjoint split
        (percentages; mean$\pm$std over three seeds).}
    \label{tab:main_loss_ablation}
\end{table}

\begin{table}[!htbp]
    \centering
    \scriptsize
    \setlength{\tabcolsep}{3.0pt}
    \renewcommand{\arraystretch}{1.08}
    \resizebox{\columnwidth}{!}{%
        \begin{tabular}{lcccccc}
            \toprule
            $\tau$ & R@1 & R@5 & R@10 & R@20 & MRR & mAP \\
            \midrule
            0.01 & 0.087{\scriptsize$\pm$0.009} & 0.370{\scriptsize$\pm$0.027} & 0.777{\scriptsize$\pm$0.062} & 1.350{\scriptsize$\pm$0.124} & 0.465{\scriptsize$\pm$0.007} & 0.387{\scriptsize$\pm$0.024} \\
            0.05 & \textbf{0.180}{\scriptsize$\pm$0.058} & \textbf{0.486}{\scriptsize$\pm$0.034} & \textbf{0.936}{\scriptsize$\pm$0.085} & \textbf{1.674}{\scriptsize$\pm$0.115} & \textbf{0.614}{\scriptsize$\pm$0.025} & \textbf{0.486}{\scriptsize$\pm$0.036} \\
            0.10 & 0.098{\scriptsize$\pm$0.035} & 0.322{\scriptsize$\pm$0.066} & 0.746{\scriptsize$\pm$0.103} & 1.394{\scriptsize$\pm$0.180} & 0.465{\scriptsize$\pm$0.090} & 0.370{\scriptsize$\pm$0.066} \\
            0.20 & 0.123{\scriptsize$\pm$0.041} & 0.396{\scriptsize$\pm$0.085} & 0.671{\scriptsize$\pm$0.027} & 1.168{\scriptsize$\pm$0.120} & 0.471{\scriptsize$\pm$0.003} & 0.367{\scriptsize$\pm$0.063} \\
            \bottomrule
        \end{tabular}%
    }
    \caption{EEG SupCon reconstruction-assisted Transformer temperature sweep on the
        episode-disjoint split (percentages; mean$\pm$std over three seeds).}
    \label{tab:main_temperature}
\end{table}

\subsection{Correspondence controls and selective held-out-training checks}
Table~\ref{tab:sanity_loso_controls} reports the correspondence controls, and
Table~\ref{tab:loso_multi} reports three-seed held-out-training checks for
fMRI--Omni and Joint-Dual--Omni with subjects sub-0001 and sub-0002 excluded
from training.  The label-aware random-brain and one-to-one shuffled-gallery
controls remain at their respective retrieval floors under both anchors,
supporting the neural--visual correspondence measured by the learned systems.

\begin{table}[!htbp]
    \centering
    \scriptsize
    \setlength{\tabcolsep}{4.0pt}
    \renewcommand{\arraystretch}{1.10}
    \resizebox{\columnwidth}{!}{%
        \begin{tabular}{llccccc}
            \toprule
            Setting                        & Split / pool  & s2v@1 & v2s@1 & Mean R@1 & MRR   & nDCG@10 \\
            \midrule
            Label-aware theoretical chance & 6480          & 0.026 & 0.026 & 0.026    & 0.222 & 0.081   \\
            Random brain, Omni             & 6480          & 0.015 & 0.015 & 0.015    & 0.149 & 0.052   \\
            Shuffled video, Omni           & 6480          & 0.000 & 0.000 & 0.000    & 0.125 & 0.045   \\
            Random brain, ImageBind        & 6480          & 0.015 & 0.000 & 0.008    & 0.151 & 0.052   \\
            Shuffled video, ImageBind      & 6480          & 0.000 & 0.000 & 0.000    & 0.129 & 0.054   \\
            \bottomrule
        \end{tabular}
    }
    \caption{Retrieval controls (percentages).  Theoretical chance and random-brain
        rows are label-aware; shuffled-video rows use one-to-one row correspondence
        after permuting the video gallery.}
    \label{tab:sanity_loso_controls}
\end{table}

\begin{table}[!htbp]
    \centering
    \scriptsize
    \setlength{\tabcolsep}{3.0pt}
    \renewcommand{\arraystretch}{1.08}
    \resizebox{\columnwidth}{!}{%
        \begin{tabular}{llcccccc}
            \toprule
            Model & Subject & R@1 & R@5 & R@10 & R@20 & MRR & mAP \\
            \midrule
            fMRI--Omni  & sub-0001 & 5.234{\scriptsize$\pm$5.239} & 16.501{\scriptsize$\pm$13.148} & 24.103{\scriptsize$\pm$15.745} & 31.956{\scriptsize$\pm$14.967} & 11.083{\scriptsize$\pm$8.145} & 11.085{\scriptsize$\pm$8.146} \\
            fMRI--Omni  & sub-0002 & 0.050{\scriptsize$\pm$0.037} & 0.197{\scriptsize$\pm$0.053}   & 0.367{\scriptsize$\pm$0.093}    & 0.768{\scriptsize$\pm$0.157}    & 0.288{\scriptsize$\pm$0.061}  & 0.288{\scriptsize$\pm$0.061} \\
            J-Dual--Omni & sub-0001 & \textbf{24.150}{\scriptsize$\pm$1.944} & \textbf{42.934}{\scriptsize$\pm$1.379} & \textbf{47.909}{\scriptsize$\pm$0.725} & \textbf{49.792}{\scriptsize$\pm$0.241} & \textbf{32.203}{\scriptsize$\pm$1.652} & \textbf{32.203}{\scriptsize$\pm$1.653} \\
            J-Dual--Omni & sub-0002 & 0.069{\scriptsize$\pm$0.012} & 0.274{\scriptsize$\pm$0.059}   & 0.505{\scriptsize$\pm$0.120}    & 0.860{\scriptsize$\pm$0.194}    & 0.320{\scriptsize$\pm$0.019}  & 0.320{\scriptsize$\pm$0.019} \\
            \bottomrule
        \end{tabular}%
    }
    \caption{Selective three-seed held-out-training checks (percentages; mean$\pm$std
        over seeds).  Twenty percent of each held-out subject's samples are used for
        checkpoint selection; the remaining samples form the test gallery.}
    \label{tab:loso_multi}
\end{table}

The held-out-training results are strongly subject-dependent.  Joint-Dual--Omni
is consistently strong for sub-0001 ($24.150\pm1.944\%$ mean R@1) but remains
close to the floor for sub-0002 ($0.069\pm0.012\%$).  fMRI--Omni shows the same
direction of subject shift, with much higher seed variance on sub-0001
($5.234\pm5.239\%$).  This selective matrix covers only two subjects and two
Omni models, and checkpoint selection still uses a validation subset from the
held-out participant.  It remains a held-out-training check and provides no
population-level cross-subject estimate.

\subsection{Low-capacity probes calibrate frozen-proxy readability}
Shallow linear and MLP probes map EEG or fMRI features directly into the same
frozen video proxy spaces under the random split using a 2{,}000-candidate
diagnostic gallery.
Table~\ref{tab:main_probe_selection} shows that Omni readouts exceed ImageBind
readouts, and fMRI readouts exceed EEG readouts within this probe protocol.

\begin{table*}[t]
    \centering
    \scriptsize
    \setlength{\tabcolsep}{4.2pt}
    \renewcommand{\arraystretch}{1.12}
    \resizebox{\textwidth}{!}{%
        \begin{tabular}{lllccccccc}
            \toprule
            Anchor    & Input & Probe  & R@1 & s2v@1 & v2s@1 & s2v@5 & v2s@5 & s2v@10 & v2s@10 \\
            \midrule
            ImageBind & EEG   & Linear & 0.55     & 0.50  & 0.60  & 1.15  & 1.90  & 2.05   & 3.80   \\
            Omni      & EEG   & Linear & 1.05     & 0.95  & 1.15  & 3.20  & 3.65  & 6.20   & 5.60   \\
            ImageBind & EEG   & MLP    & 0.80     & 0.80  & 0.80  & 2.45  & 2.30  & 3.00   & 4.00   \\
            Omni      & EEG   & MLP    & 1.23     & 1.15  & 1.30  & 4.05  & 4.55  & 6.80   & 7.35   \\
            ImageBind & fMRI  & MLP    & 18.58    & 17.20 & 19.95 & 27.75 & 35.45 & 34.30  & 43.15  \\
            Omni      & fMRI  & MLP    & 22.85    & 20.60 & 25.10 & 47.25 & 52.30 & 58.40  & 64.95  \\
            \bottomrule
        \end{tabular}%
    }
    \caption{Low-capacity probe results on a 2{,}000-candidate CineBrain random-split gallery (percentages).
        Linear and MLP probes use shallow projections into the frozen video proxy
        space and the same label-aware bidirectional retrieval protocol as BrainBind.}
    \label{tab:main_probe_selection}
\end{table*}

The probes reproduce the main anchor and modality trends: Omni is more readable
than ImageBind for EEG (linear 1.05 vs.\ 0.55; MLP 1.23 vs.\ 0.80), and the
fMRI MLP probe reaches 22.85 mean R@1 with Omni versus 18.58 with ImageBind.
These within-gallery results show that the frozen proxies are readily accessible
to shallow neural readouts.  A separate 6{,}480-candidate EEG--Omni evaluation
in the supplementary material places the linear and MLP probes at 0.32 and
0.40 mean R@1, versus 1.91 for the full encoder under the same gallery size.

\subsection{Episode-disjoint evaluation exposes context dependence}
As a complementary generalization test, we group all clips by
season--episode identifier and assign every
clip from an episode to exactly one split.  The resulting 30 disjoint episodes
contain 19{,}444/6{,}480/6{,}480 train/validation/test sample pairs.  Retrieval drops
sharply under this stricter protocol (Table~\ref{tab:episode_disjoint}), showing
that repeated stimulus identity and episode continuity contribute materially
to random-split retrieval.  The label-aware R@1 chance floor is
0.069\%: EEG is close to this floor, fMRI is higher, and Joint-Dual has the
largest mean at every reported $K$.  Each of the three runs couples its model
seed to the episode-assignment seed, capturing both initialization and split
variation in the reported descriptive statistics.

\begin{table}[!htbp]
    \centering
    \scriptsize
    \setlength{\tabcolsep}{3.2pt}
    \renewcommand{\arraystretch}{1.10}
    \resizebox{0.85\columnwidth}{!}{%
    \begin{tabular}{lccccc}
        \toprule
        Model & $n$ & R@1 & R@5 & R@10 & R@20 \\
        \midrule
        \emph{Chance floor} & -- & 0.069 & 0.347 & 0.693 & 1.381 \\
        \midrule
        EEG    & 3 & 0.075$\pm$0.029 & 0.363$\pm$0.109 & 0.579$\pm$0.182 & 1.224$\pm$0.323 \\
        fMRI   & 3 & 0.129$\pm$0.073 & 0.417$\pm$0.250 & 0.908$\pm$0.244 & 1.641$\pm$0.373 \\
        Both   & 3 & 0.126$\pm$0.093 & 0.453$\pm$0.200 & 0.893$\pm$0.462 & 1.402$\pm$0.585 \\
        J-Dual & 3 & \textbf{0.216}$\pm$0.051 & \textbf{0.615}$\pm$0.083 & \textbf{1.026}$\pm$0.109 & \textbf{1.988}$\pm$0.054 \\
        \bottomrule
    \end{tabular}
    }
    \caption{Episode-disjoint label-aware retrieval (percentages;
        mean$\pm$std over three coupled model/split-seed runs).  No season--episode group appears in more
        than one split; the chance row reflects the repeated-positive structure.}
    \label{tab:episode_disjoint}
\end{table}

\subsection{Retrieval-conditioned visualization provides qualitative examples}
\label{sec:proxy_generation}

As a focused interpretability aid, the fMRI--Omni embedding retrieves a text
description.  A prompt-only diagnostic uses the first 32 test-loader samples as
both queries and gallery candidates (chance $1/32=3.1\%$).  In a separate
generation pass, retrieved text from a 16-query run conditions the fixed
Wan2.1-T2V-1.3B generator \cite{wan2025}; ten outputs were retained.  Figure~\ref{fig:overview}
and the supplementary show representative generated frames from this
proxy-mediated pathway, not direct neural reconstruction.

\section{Discussion}
\label{sec:discussion}

\paragraph{Video proxy geometry as a shared neural reference.}
Within CineBrain, frozen long-form video embeddings provide a common retrieval
geometry for synchronized EEG and fMRI.  Their role here is a stimulus
reference.  The 12.32\,pp fMRI nDCG@10 advantage of Omni over
ImageBind ($p=1.3\times10^{-6}$) coincides with Omni's broader effective
spectrum (effective rank 1{,}112 vs.\ 426) and lower redundancy.  The association
is observational.  The consistent Omni advantage across all eight systems shows
that proxy choice is associated with alignment quality.  Selecting and characterizing the target
video space is therefore a central design decision in neural alignment.

\paragraph{Modality- and system-specific profiles.}
The modalities and systems expose different retrieval profiles.  fMRI-only
provides the strongest rank-wide organization, while EEG-only carries a
measurable, distinct in-domain retrieval profile alongside fMRI's stronger
rank-wide organization.
The temporal-order perturbation result suggests that the current EEG encoder
prioritizes order-robust features.  Joint-Dual places a label-positive candidate
first more often than fMRI-only as a complete S4AE, capacity, and multi-branch
system.  Its rank-wide metrics make system designs that combine top-rank
precision with broader candidate organization a promising direction.

\paragraph{Relation to static-image benchmarks.}


BrainBind evaluates long-form synchronized video, complementing established paradigms such as THINGS-EEG and the NSD with MindEye, which primarily rely on isolated, static images or brief trial epochs. While static benchmarks isolate fine-grained visual representations without temporal confounding, continuous naturalistic viewing introduces complex narrative flow, multi-scale temporal dependencies, and recurring semantic contexts that fundamentally alter the neural decoding landscape. Direct cross-benchmark score comparisons can therefore be misleading due to structural differences in stimulus density,  continuous hemodynamic integration, and evaluation dynamics. Consequently, we position our baseline controls, ranking-aware metrics, and visual embedding diagnostics strictly within the naturalistic video protocol on CineBrain, offering a calibrated framework for evaluating synchronized multimodal brain representations under dynamic real-world conditions.

\paragraph{Implications for evaluation.}
The results also clarify how synchronized neural decoding should be evaluated.
The random split measures fine-grained in-distribution retrieval with repeated
stimulus identity and narrative context, while the episode-disjoint split
complements it with unseen-episode evaluation.  Reporting top-rank and rank-wide
metrics under both protocols reveals precision and candidate-space organization
within a unified evaluation framework.

\paragraph{Scope and future directions.}


The current CineBrain study establishes a synchronized long-form video benchmark evaluated under two complementary splitting protocols. Under the four-observation 
structure, approximately $93.6\%$ of random-split test samples share exact segment identities  with the training set, capturing model performance when exposed to recurring dynamic scenes. Conversely, the episode-disjoint evaluation strips away this overlap to quantify true out-of-domain narrative generalization. Initial held-out checks across two subjects and three random seeds provide an early snapshot of inter-subject variability, while hyperparameter ablations over contrastive losses 
(e.g., SupCon) and temperature parameters optimize representation learning within specific model families. Furthermore, our qualitative generation panel converts neural retrieval embeddings into descriptive text prompts to condition generative video models (such as Wan2.1), creating an interpretable semantic bridge alongside quantitative retrieval matrices. Future work will build upon this foundation by implementing parameter-matched fMRI controls to isolate modality-specific representation capacity, scaling to complete cross-subject folds, and integrating multi-positive contrastive objectives to accommodate non-unique semantic matches in continuous video streams.

\section{Conclusion}
\label{sec:conclusion}

We introduced BrainBind, a stimulus-aligned framework that compares synchronized
EEG, fMRI, and fused systems within frozen long-form video geometries.  On
CineBrain, the $2\!\times\!4$ system matrix shows:
fMRI--Omni provides the strongest rank-wide organization, EEG--Omni remains
above controls, and Joint-Dual--Omni most often places a label-positive
candidate first with a distinct rank-wide profile.  Omni outperforms
ImageBind throughout the matrix, linking proxy choice to alignment quality, and
the episode-disjoint drop reveals the contribution of recurring narrative
 context to random-split scores.  Episode-disjoint loss and temperature
 ablations further identify SupCon and $\tau{=}0.05$ as stronger EEG
 configurations under this stricter protocol, while two-subject held-out checks underscore the subject dependence
 that remains to be characterized.  Correspondence controls and geometry
 diagnostics support the observed synchronized neural--video alignment within
 the scope of the present dataset. The current evidence points to three concrete next steps: a capacity-matched
fMRI-only control, complete subject folds, and multi-positive objectives for
related clips.  These tests would separate multimodal information from capacity,
estimate population-level subject variability, and align the objective with the
repeated-label structure of naturalistic video.

    {\small
        \setlength{\bibsep}{0pt}
        \bibliography{brainbind_refs}
    }

\newpage
\clearpage

\section*{BrainBind: Supplementary Material}

\section*{Overview}

This supplement provides extended protocol details, S4AE configuration results,
correspondence controls, held-out low-capacity probes, cross-anchor geometry,
qualitative retrieval-conditioned visualization, selective held-out-training
checks, and hardware details that support the main paper.  The primary
retrieval, ranking, statistical, ablation, probe, and training tables are
reported in the main text to avoid duplication.

\section{Experimental Protocol}

This section specifies the data alignment, split construction, and retrieval protocol used for all reported experiments.

\subsection{Dataset and Temporal Alignment}

All experiments use CineBrain~\cite{cinebrain}, a synchronized EEG--fMRI
dataset collected during naturalistic audiovisual viewing. Each sample
pairs a video segment with its corresponding EEG and fMRI windows under
the dataset-provided temporal alignment protocol, including
hemodynamic-response compensation for fMRI windows. This keeps every model
variant under a common temporal reference and enables configuration-level comparisons
across modality, proxy anchor, fusion implementation, and backbone.

BrainBind treats frozen visual representations as \emph{proxy stimulus
    geometries}: structured anchors for neural alignment, not literal neural
codes. The proxy space supplies a
structured visual-semantic reference within which heterogeneous neural
measurements can be associated with complex video scenes.  It evaluates how EEG
and fMRI responses project into a shared video-grounded space, using the proxy
embedding as a stable stimulus reference for neural alignment.

\subsection{Evaluation Splits}

The main retest matrix uses a fixed 60/20/20 random split.  All configurations
use identical partitions, enabling configuration-level comparisons across
neural modality, proxy anchor, fusion implementation, and backbone.
With approximately four observations per stimulus, the expected probability
that a random-split test sample shares its exact segment with training is
$1-0.4^3=93.6\%$.  This split provides an in-distribution retrieval setting,
complemented by the episode-disjoint evaluation below.

We further report selective held-out-training checks for fMRI--Omni and
Joint-Dual--Omni with subjects sub-0001 and sub-0002 excluded from training.
For each held-out subject, 20\% of that participant's samples are used for
validation and the remaining 80\% for test, with seeds 42, 123, and 456.  This
asks whether video-grounded alignment remains measurable when a participant is
absent from the training set.  The two reported subjects provide an initial
view of subject variation and motivate complete LOSO folds.

We additionally use an episode-disjoint split to test sensitivity to repeated
narrative context.  All clips sharing a season--episode identifier are assigned
to exactly one partition.  The resulting 30 disjoint episodes contain
19{,}444/6{,}480/6{,}480 train/validation/test samples, so no episode appears in
more than one split.  The three runs couple model and episode-assignment seeds
42, 123, and 456, incorporating both initialization and split variation.

\subsection{Retrieval Evaluation}

Retrieval is evaluated over the full evaluation split.  In the label-aware
setting, all candidates with the same \texttt{stim\_label} are positives, and
candidates with different \texttt{stim\_label} values are negatives.
Brain-to-video retrieval (s2v) uses neural embeddings as queries against the
video proxy gallery; video-to-brain retrieval (v2s) reverses the direction.

All metrics (R@1, R@5, R@10, R@20, mean R@K, MRR, mAP, and nDCG@10) are
computed from the ranked list for each query direction.  This unified protocol allows the
evaluation to capture both top-rank matching and the global organization of the
candidate space.  R@1 measures whether any label-positive candidate appears
first, while MRR, mAP, nDCG, and higher-rank recall describe how positives are
structured throughout the retrieval list.
Core Recall aggregates use five seeds.  EEG--ImageBind MRR, mAP, and nDCG@10
use four available seeds because one run lacks ranking outputs; every other
core ranking aggregate uses five.

\section{S4AE Architecture and Targeted Configuration Diagnostics}
\label{sec:supp_s4ae_design}

S4AE denotes the backbone family; reconstruction is enabled only in designated
diagnostic configurations.  The core
Joint-Dual systems already contain independent EEG and fMRI S4AE branches and
use contrastive-only training.  We additionally report a reduced-width
Joint-Dual retest and an EEG--ImageBind S4AE configuration.  These targeted rows
differ in width and objective and extend the primary eight-system matrix without
completing a matched factorial comparison.

\paragraph{Architecture.}
Given an EEG or fMRI window, a lightweight TemporalMLP maps the input to the
model width before stacked structured state-space and channel-mixing blocks.  Normalization
and residual connections wrap both mixers, and the aggregated bottleneck is
projected into the selected proxy space.  For EEG, the structured mixer operates
along sampled time.  In the reported fMRI branch, the five TRs are input
channels and the 18{,}926 voxels form the mixed sequence axis; we therefore use
the neutral term \emph{structured state-space mixing} for fMRI.  When
reconstruction is enabled, mirrored decoder blocks map the latent sequence back
to the active neural input.

\paragraph{Objective.}
The optional joint objective for an active modality $m$ is
\begin{equation}
    \mathcal{L}_{\mathrm{S4AE}}
    =\beta\mathcal{L}_{\mathrm{NCE}}+\alpha\mathcal{L}_{\mathrm{rec}},
    \quad
    \mathcal{L}_{\mathrm{rec}}
    =\frac{1}{|\mathbf{X}^{m}|}\|\hat{\mathbf{X}}^{m}-\mathbf{X}^{m}\|_1 .
\end{equation}
All reported configurations set $\beta=1$.  The core Joint-Dual and reduced-width Joint-Dual retest set $\alpha=0$ and have
no reconstruction loss.  The EEG--ImageBind diagnostic uses joint contrastive
and reconstruction training with $\alpha=0.25$.  In the SupCon ablations,
positives share the same \texttt{stim\_label}; the Transformer decoder predicts
the mean-pooled 512-dimensional projected EEG input.

\paragraph{Interpretation.}
The S4AE rows are configuration diagnostics that reveal useful architecture,
width, and objective choices and motivate a fuller modality-by-anchor grid.

\section{Additional S4AE Configuration Results}

Tables~\ref{tab:v1_new_ablation_matrix} and
\ref{tab:v1_new_ablation_ranking} report the reduced-width Joint-Dual--Omni
configuration over three seeds and the EEG--ImageBind S4AE configuration over
five seeds.  They remain separate from the primary matrix to present targeted
evidence over backbone, width, objective, modality, and anchor choices.

\providecommand{\val}[2]{#1$\pm$#2}

\begin{table}[!htbp]
    \centering
    \small
    \setlength{\tabcolsep}{4pt}
    \renewcommand{\arraystretch}{1.15}
    \begin{tabular}{llccccc}
    \toprule
    Anchor / input & Direction & $n$ & R@1 & R@5 & R@10 & R@20 \\
    \midrule
    Omni / J-Dual & s2v & 3 & \val{10.08}{0.32} & \val{17.44}{0.51} & \val{22.96}{1.13} & \val{30.92}{1.25} \\
    Omni / J-Dual & v2s & 3 & \val{18.68}{0.42} & \val{27.14}{0.81} & \val{35.06}{1.47} & \val{44.98}{2.04} \\
    \midrule
    ImageBind / EEG & s2v & 5 & \val{0.09}{0.06} & \val{0.30}{0.09} & \val{0.51}{0.10} & \val{0.95}{0.18} \\
    ImageBind / EEG & v2s & 5 & \val{0.20}{0.03} & \val{0.82}{0.07} & \val{1.49}{0.11} & \val{2.69}{0.30} \\
    \bottomrule
    \end{tabular}
    \caption{Targeted S4AE configurations on CineBrain bidirectional retrieval.
        Values are percentages; $n$ is the number of seeds. The reduced-width
        J-Dual configuration uses mean fusion and contrastive-only training.
        The single-modality EEG--ImageBind configuration jointly optimizes
        contrastive alignment and reconstruction. Widths, objectives, anchors,
        and seed sets differ, so these are configuration diagnostics rather
        than a factorial backbone ablation.}
    \label{tab:v1_new_ablation_matrix}
\end{table}

\begin{table}[!htbp]
    \centering
    \scriptsize
    \setlength{\tabcolsep}{4.0pt}
    \renewcommand{\arraystretch}{1.12}
    \caption{Ranking-aware metrics for the targeted S4AE configurations. Values
        are percentages, mean$\pm$std over $n$ seeds.}
    \label{tab:v1_new_ablation_ranking}
    \resizebox{\columnwidth}{!}{%
        \begin{tabular}{llllccccc}
            \toprule
            Anchor    & Input  & Fusion & Configuration             & $n$ & Mean R@1          & MRR               & mAP               & nDCG@10           \\
            \midrule
            Omni      & J-Dual & mean   & reduced S4AE, contrastive & 3   & \val{14.38}{0.35} & \val{19.14}{0.56} & \val{15.28}{0.52} & \val{16.89}{0.66} \\
            ImageBind & EEG    & single & S4AE + reconstruction     & 5   & \val{0.15}{0.03}  & \val{0.59}{0.05}  & \val{0.50}{0.04}  & \val{0.39}{0.04}  \\
            \bottomrule
        \end{tabular}
    }
\end{table}

\paragraph{Takeaway.} With Omni, the reduced-width contrastive-only Joint-Dual
model reaches mean R@1 14.38 over three seeds, compared with 13.52 for the larger
five-seed core system.  Joint S4AE training with EEG--ImageBind reaches 0.15,
below the Transformer row at 0.61.  These results provide useful capacity and
configuration checks; the differing seed sets and objectives support
configuration-level rather than single-factor interpretation.

\section{Chance and Shuffled-Feature Controls}
\label{sec:supp_controls}

The control experiments evaluate retrieval when neural--video correspondence is
removed or replaced by theoretical chance structure.  All values are reported as
percentages.  These controls are important because video embeddings can contain
intrinsic clustering structure; therefore, a valid neural--video retrieval
signal should remain separated from random-brain and shuffled-video baselines.
The shuffled-video rows use unshuffled frozen video features as queries and a
seed-fixed permutation of the same features as the gallery, with exact row
identity as the positive definition; they do not use trained neural embeddings.

\subsection{Omni Controls}
\label{sec:supp_omni_controls}

Table~\ref{tab:supp_omni_controls} reports the Omni control results.  The
one-to-one theoretical chance row corresponds to exact pair matching in the full
candidate pool.  The label-aware theoretical chance row accounts for repeated
same-\texttt{stim\_label} positives.  Random-brain and shuffled-video controls
 remain at their respective retrieval floors, providing label-aware and
proxy-only empirical references for the learned neural--video results.

\begin{table}[!htbp]
    \centering
    \scriptsize
    \resizebox{\columnwidth}{!}{%
        \begin{tabular}{lcccccccc}
            \toprule
            Control                        & s2v@1 & s2v@5 & s2v@10 & s2v@20 & v2s@1 & v2s@5 & v2s@10 & v2s@20 \\
            \midrule
            One-to-one theoretical chance  & 0.015 & 0.077 & 0.154  & 0.309  & 0.015 & 0.077 & 0.154  & 0.309  \\
            Label-aware theoretical chance & 0.026 & 0.131 & 0.262  & 0.523  & 0.026 & 0.131 & 0.262  & 0.523  \\
            Label-aware random brain       & 0.015 & 0.046 & 0.077  & 0.123  & 0.015 & 0.093 & 0.139  & 0.401  \\
            Shuffled video features        & 0.000 & 0.062 & 0.108  & 0.262  & 0.000 & 0.046 & 0.123  & 0.355  \\
            \bottomrule
        \end{tabular}
    }
    \caption{Omni chance and shuffled-feature controls. Label-aware rows use
        same-\texttt{stim\_label} positives; the shuffled-video row reports the
        one-to-one shuffled-feature control.}
    \label{tab:supp_omni_controls}
\end{table}

\subsection{ImageBind Controls}
\label{sec:supp_imagebind_controls}

Table~\ref{tab:supp_imagebind_controls} reports the corresponding ImageBind
controls.  As with Omni, random-brain and shuffled-video rows remain at the
retrieval floor.  The learned ImageBind systems remain clearly separated from
both the label-aware random-brain and one-to-one proxy-only references.

\begin{table}[!htbp]
    \centering
    \scriptsize
    \resizebox{\columnwidth}{!}{%
        \begin{tabular}{lcccccccc}
            \toprule
            Control                        & s2v@1 & s2v@5 & s2v@10 & s2v@20 & v2s@1 & v2s@5 & v2s@10 & v2s@20 \\
            \midrule
            One-to-one theoretical chance  & 0.015 & 0.077 & 0.154  & 0.309  & 0.015 & 0.077 & 0.154  & 0.309  \\
            Label-aware theoretical chance & 0.026 & 0.131 & 0.262  & 0.523  & 0.026 & 0.131 & 0.262  & 0.523  \\
            Label-aware random brain       & 0.015 & 0.062 & 0.139  & 0.278  & 0.000 & 0.077 & 0.108  & 0.231  \\
            Shuffled video features        & 0.000 & 0.031 & 0.077  & 0.293  & 0.000 & 0.108 & 0.231  & 0.309  \\
            \bottomrule
        \end{tabular}
    }
    \caption{ImageBind chance and shuffled-feature controls. Label-aware rows use
        same-\texttt{stim\_label} positives; the shuffled-video row reports the
        one-to-one shuffled-feature control.}
    \label{tab:supp_imagebind_controls}
\end{table}

\subsection{Ranking-Aware Controls}
\label{sec:supp_ranking_controls}

Table~\ref{tab:supp_ranked_controls} reports ranking-aware control metrics.
The control rows remain near the theoretical floor for both anchors, while the
main BrainBind configurations remain well above these baselines.  This validates
that the reported retrieval signal is not explained by intrinsic clustering of
video embeddings, random neural structure, or label-aware bookkeeping.

\begin{table}[!htbp]
    \centering
    \scriptsize
    \resizebox{\columnwidth}{!}{%
        \begin{tabular}{llcccc}
            \toprule
            Anchor    & Control                        & Mean R@1 & Mean MRR & Mean mAP & Mean nDCG@10 \\
            \midrule
            Omni      & Label-aware theoretical chance & 0.026    & 0.222    & 0.155    & 0.081        \\
            Omni      & Label-aware random brain       & 0.015    & 0.149    & 0.148    & 0.052        \\
            Omni      & Shuffled video features        & 0.000    & 0.125    & 0.125    & 0.045        \\
            ImageBind & Label-aware theoretical chance & 0.026    & 0.222    & 0.155    & 0.081        \\
            ImageBind & Label-aware random brain       & 0.008    & 0.151    & 0.154    & 0.052        \\
            ImageBind & Shuffled video features        & 0.000    & 0.129    & 0.129    & 0.054        \\
            \bottomrule
        \end{tabular}
    }
    \caption{Ranking-aware chance and control metrics. The label-aware theoretical
        chance accounts for multiple same-label positives per query. The shuffled-video
        rows use the one-to-one shuffled-feature control.}
    \label{tab:supp_ranked_controls}
\end{table}

\paragraph{Takeaway.} Across both anchors and both retrieval directions,
random-brain and shuffled-video controls remain at their respective retrieval
floors.  The separation between BrainBind results and these controls is large in
absolute terms (fMRI--Omni mean R@1 of 12.71 versus a label-aware floor of
0.026) and consistent across recall and ranking metrics.  The control
separation is especially informative for EEG: above-chance retrieval tracks
preserved neural--video correspondence, while randomized neural inputs remain
at floor level in the same anchor.

\section{Low-Capacity Probe Baselines}
\label{sec:supp_low_capacity_probes}

The main paper reports the low-capacity probe matrix on a CineBrain random-split
gallery with 2{,}000 candidates.  For held-out EEG with Omni, we additionally
report probe runs on the test gallery with 6{,}480 candidates, enabling a same-gallery
comparison with the full encoder.

\begin{table}[!htbp]
    \centering
    \scriptsize
    \setlength{\tabcolsep}{4.5pt}
    \renewcommand{\arraystretch}{1.12}
    \resizebox{\columnwidth}{!}{%
        \begin{tabular}{llccccccc}
            \toprule
            Input & Probe  & Mean R@1 & s2v@1 & v2s@1 & s2v@5 & v2s@5 & s2v@10 & v2s@10 \\
            \midrule
            EEG   & Linear & 0.32     & 0.29  & 0.35  & 1.02  & 1.68  & 1.87   & 2.87   \\
            EEG   & MLP    & 0.40     & 0.29  & 0.51  & 1.36  & 2.28  & 2.27   & 3.60   \\
            \bottomrule
        \end{tabular}
    }
    \caption{Held-out test results for completed EEG--Omni probe runs.  Values are
        percentages.  The probes and full EEG--Omni encoder use the same
        6{,}480-candidate test gallery and label-aware retrieval protocol.}
    \label{tab:supp_probe_test_eeg_omni}
\end{table}

\paragraph{Held-out test calibration.}
The completed EEG--Omni test runs show that shallow readouts recover measurable
video-aligned information from EEG.  The MLP
probe improves over the linear probe in mean R@1 (0.40 vs. 0.32) and
higher-rank recall, indicating that a small nonlinear readout helps extract
EEG-proxy correspondence.  The full BrainBind EEG--Omni encoder in the main
matrix reaches 1.91 mean R@1, substantially above both low-capacity probes.
This gap supports the contribution of the learned neural encoder and
contrastive proxy alignment beyond a shallow mapping from EEG features.
Alongside the random-brain and shuffled-video controls, the probe baselines show that
BrainBind's retrieval signal reflects measurable neural--video association
and not a trivial readout from proxy-only structure.

\paragraph{Takeaway.}
Low-capacity probes provide an interpretable baseline for proxy-space
readability.  They reproduce two central trends from the main experiments:
Omni is more readable than ImageBind, and fMRI exposes stronger video-aligned
structure than EEG.  At the same time, the held-out EEG--Omni comparison shows
that BrainBind's full neural encoder improves substantially over shallow
linear or MLP readouts.  These results strengthen the interpretation that
BrainBind benefits from learned neural-to-proxy alignment while preserving the
anchor and modality effects observed in the controlled retrieval matrix.

\section{Cross-Anchor Proxy-Space Characterization}
\label{sec:supp_cross_anchor}

\paragraph{Why this matters.} The main paper identifies anchor geometry
as a first-class design axis: every Omni configuration outperforms its
ImageBind counterpart, and Omni exceeds ImageBind by $12.32$\,pp in fMRI
nDCG@10. To examine the associated geometry, we characterize the two surrogate-anchor
spaces on 8{,}100 time-aligned video clip slices. For each slice, frozen
visual embeddings from ImageBind~\cite{imagebind} and
Omni~\cite{omni} retain their native 1{,}024- and 2{,}048-dimensional
representations for spectral analysis.  PCA is then fitted independently to
each anchor with 256 components before canonical correlation analysis.
These native frozen representations are also the contrastive target spaces:
the neural output mapping matches 1{,}024-D ImageBind or 2{,}048-D Omni, and
the 256-D PCA reduction is used only for the cross-anchor analysis.

\paragraph{Spectral structure.}
The top 256 principal components explain 94.9\% of ImageBind variance and
86.9\% of Omni variance. ImageBind exhibits stronger
low-dimensional concentration, indicating that a larger portion of its
feature variance is compressed into fewer dominant directions. Omni
shows a substantially higher effective rank (1111.7 vs.\ 425.7),
indicating a broader spectrum and lower redundancy, and a lower mean
absolute feature--feature correlation. This association offers a geometric
hypothesis for the empirical retrieval gap: Omni exposes a broader and less
redundant long-form video geometry that may be more readable by the neural
encoders and is consistent with its retrieval advantage.

\paragraph{Exploratory linear correspondence.}
CCA fitted after the independent PCA reductions yields an in-sample mean
correlation of 0.88 across the first 64 canonical components; the top 10 exceed
0.95 and the top 5 approach 0.98.  Because the same 8{,}100 slices are used for
fitting and evaluation, these values provide descriptive evidence of linear
cross-anchor correspondence and motivate held-out multi-anchor analysis.

\paragraph{Takeaway.} The geometric results contextualize the
main paper's anchor finding. Omni's broader spectrum and lower redundancy make a
richer geometry available to neural alignment and are consistent with the Omni
advantage across the evaluated configurations.  The in-sample CCA fit further
shows that a strong linear correspondence can be estimated between the two
anchors.  BrainBind therefore treats anchor choice as an explicit experimental
variable and motivates held-out multi-anchor fusion as a natural extension.

\section{Qualitative Retrieval-Conditioned Visualization}
\label{sec:supp_qualitative_decoding}

\begin{figure*}[t]
    \centering
    \includegraphics[clip,width=\textwidth]{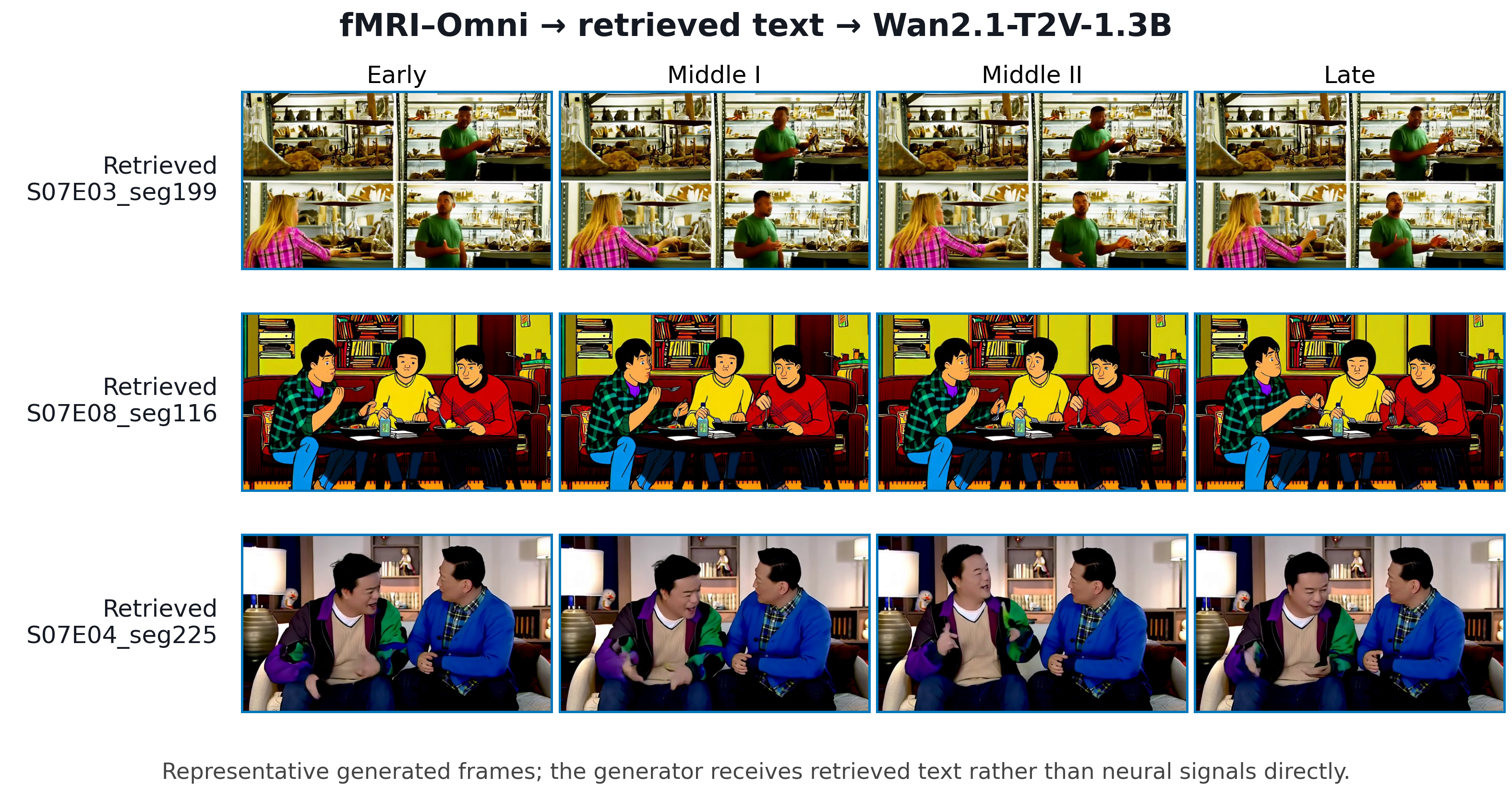}
    \caption{\textbf{Qualitative retrieval-conditioned visualization via the Omni pipeline.}
        The fMRI embedding retrieves text in Omni space, and the fixed
        Wan2.1-T2V-1.3B generator~\cite{wan2025} produces the displayed frames.
        The panel shows three of ten retained outputs from the 16-query generation
        pass.  It visualizes a neural$\rightarrow$proxy$\rightarrow$text$\rightarrow$video
        pathway rather than direct neural reconstruction.}
    \label{fig:supp_omni_pipeline}
\end{figure*}

In addition to retrieval metrics, we provide retrieval-conditioned
visualizations as an exploratory interpretability aid.  The prompt-only
diagnostic uses the first 32 test-loader samples as both queries and gallery
candidates.  A separate 16-query generation pass retrieves text in Omni space
and supplies it to Wan2.1-T2V-1.3B; ten generated videos were retained, and
Figure~\ref{fig:supp_omni_pipeline} shows representative frames from three.
Because the fixed generator receives retrieved text rather than neural signals
directly, these outputs are proxy-mediated semantic visualizations rather than
direct brain reconstruction.

\subsection{Omni pipeline: neural $\rightarrow$ proxy $\rightarrow$ text $\rightarrow$ Wan2.1}
\label{sec:supp_qual_omni}

Given a time-aligned neural window, we encode EEG and/or fMRI through the
BrainBind neural encoder to obtain an embedding in the Omni proxy space. We then
retrieve a short textual description from the same proxy space and use the
retrieved text as the prompt for Wan2.1-T2V-1.3B~\cite{wan2025}:
\begin{equation}
    \mathrm{Neural}
    \;\rightarrow\;
    \mathrm{Omni}
    \;\rightarrow\;
    \mathrm{Text}
    \;\rightarrow\;
    \mathrm{Wan2.1}.
\end{equation}
All downstream components are fixed, yielding an interpretable pathway from
neural activity through proxy embeddings to semantic descriptions.  Retrieval and generation remain separate fixed stages.

The qualitative examples show visual content associated with the retrieved
descriptions, including indoor environments, conversation settings, object
context, and human presence.  They illustrate the semantic pathway exposed by
retrieval-conditioned generation.

\section{Relation to Static-Image Neural Decoding Benchmarks}

Long-form naturalistic video complements static-image benchmarks through a
different candidate pool, stimulus regime, and target definition.  THINGS-EEG and NSD/MindEye
protocols~\cite{thingsdata,nsd,mindeye,niceeeg,atm} typically use isolated
images and short stimulus epochs, whereas BrainBind uses synchronized EEG and
fMRI windows from long-form videos with semantically proximate candidates.  We
therefore report BrainBind's absolute metrics with random-brain and
shuffled-video controls.  Chance multipliers across these incommensurate
protocols would not support a valid comparison.

\begin{table*}[t]
    \centering
    \footnotesize
    \setlength{\tabcolsep}{3.0pt}
    \renewcommand{\arraystretch}{1.18}
    \resizebox{\textwidth}{!}{%
        \begin{tabular}{>{\raggedright\arraybackslash}p{0.15\textwidth}>{\raggedright\arraybackslash}p{0.09\textwidth}>{\raggedright\arraybackslash}p{0.17\textwidth}>{\raggedright\arraybackslash}p{0.22\textwidth}>{\raggedright\arraybackslash}p{0.30\textwidth}}
            \toprule
            Study / family                                                   & Signal                                                                                                                & Stimulus / supervision                          & Output target & Scale interpretation \\
            \midrule
            MindEye2~\cite{mindeye2}                                         & fMRI                                                                                                                  & NSD static natural images
                                                                             & Image retrieval and diffusion reconstruction
                                                                             & Strong low-data static-image reference; the target is an image embedding/generator, not video dynamics.                                                                                                        \\
            MinD-Video~\cite{mindvideo}                                      & fMRI                                                                                                                  & Dynamic visual stimuli
                                                                             & Video reconstruction
                                                                             & Direct fMRI-to-video generation reference; it does not study synchronized EEG--fMRI proxy retrieval.                                                                                                           \\
            NeuroClips~\cite{neuroclips}                                     & fMRI                                                                                                                  & Natural video clips
                                                                             & Smooth fMRI-to-video reconstruction
                                                                             & High-fidelity video-generation reference optimized for perceptual synthesis, not full-split ranking.                                                                                                           \\
            Brain Netflix~\cite{brainnetflix}                                & fMRI                                                                                                                  & Scaled natural video viewing
                                                                             & Video reconstruction from brain activity
                                                                             & Closest large-scale video-generation reference; it focuses on fMRI synthesis and data scaling.                                                                                                                 \\
            Animate Your Thoughts / Mind-Animator~\cite{animateyourthoughts} & fMRI                                                                                                                  & Dynamic natural vision
                                                                             & Decoupled video reconstruction
                                                                             & Recent fMRI-to-video generation reference; its target is reconstruction, not shared EEG--fMRI retrieval geometry.                                                                                              \\
            \textbf{BrainBind (ours)}                                        & \textbf{EEG + fMRI}                                                                                                   & \textbf{CineBrain long-form naturalistic video}
                                                                             & \textbf{Bidirectional retrieval, proxy diagnostics, and retrieval-conditioned visualization}
                                                                             & \textbf{Studies whether synchronized EEG and fMRI can be organized in one frozen long-form video reference geometry.}                                                                                          \\
            \bottomrule
        \end{tabular}
    }
    \caption{Video-generation and static-image external references as scale context.
        These studies are not direct leaderboard comparisons: they differ in stimulus
        regime, signal modality, and output target. They define the landscape against
        which BrainBind's distinct contribution is positioned.}
    \label{tab:supp_video_generation_baselines}
\end{table*}

\FloatBarrier
\section{Selective Held-Out-Training Checks}
\label{sec:supp_loso}

Table~\ref{tab:supp_loso} reports three-seed held-out-training checks for
fMRI--Omni and Joint-Dual--Omni with subjects sub-0001 and sub-0002 excluded
from training.  For each held-out subject, 20\% of that participant's samples
are used for checkpoint selection and the remaining 80\% form the test gallery.
These selective checks provide a focused two-participant view of subject
variation and complement the random- and episode-split evaluations; complete
LOSO folds are a direct extension.

\begin{table}[!htbp]
    \centering
    \scriptsize
    \setlength{\tabcolsep}{4.0pt}
    \renewcommand{\arraystretch}{1.08}
    \resizebox{\textwidth}{!}{%
        \begin{tabular}{llcccccccc}
            \toprule
            Model & Held-out subject & Seed & Pairs & Mean R@1 & R@5 & R@10 & R@20 & MRR & mAP \\
            \midrule
            fMRI--Omni  & sub-0001 & 42  & 4321 & 11.247 & 31.486 & 41.865 & 48.554 & 20.368 & 20.372 \\
            fMRI--Omni  & sub-0001 & 123 & 4321 & 2.800  & 11.120 & 18.584 & 27.829 & 7.737  & 7.738  \\
            fMRI--Omni  & sub-0001 & 456 & 4321 & 1.655  & 6.897  & 11.861 & 19.486 & 5.145  & 5.145  \\
            fMRI--Omni  & sub-0002 & 42  & 4320 & 0.035  & 0.185  & 0.359  & 0.671  & 0.270  & 0.270  \\
            fMRI--Omni  & sub-0002 & 123 & 4320 & 0.093  & 0.255  & 0.463  & 0.949  & 0.356  & 0.356  \\
            fMRI--Omni  & sub-0002 & 456 & 4320 & 0.023  & 0.150  & 0.278  & 0.683  & 0.237  & 0.237  \\
            \midrule
            J-Dual--Omni & sub-0001 & 42  & 4321 & 21.951 & 41.356 & 47.072 & 49.514 & 30.314 & 30.314 \\
            J-Dual--Omni & sub-0001 & 123 & 4321 & 24.855 & 43.532 & 48.322 & 49.931 & 32.914 & 32.914 \\
            J-Dual--Omni & sub-0001 & 456 & 4321 & 25.642 & 43.913 & 48.334 & 49.931 & 33.380 & 33.382 \\
            J-Dual--Omni & sub-0002 & 42  & 4320 & 0.058  & 0.324  & 0.602  & 0.984  & 0.342  & 0.342  \\
            J-Dual--Omni & sub-0002 & 123 & 4320 & 0.069  & 0.289  & 0.544  & 0.961  & 0.315  & 0.316  \\
            J-Dual--Omni & sub-0002 & 456 & 4320 & 0.081  & 0.208  & 0.370  & 0.637  & 0.304  & 0.304  \\
            \bottomrule
        \end{tabular}%
    }
    \caption{Selective held-out-training checks (percentages).  Mean R@1, R@5,
        R@10, R@20, MRR, and mAP average the neural-to-video and
        video-to-neural directions.}
    \label{tab:supp_loso}
\end{table}

Across the three seeds, Joint-Dual--Omni reaches mean R@1
$24.150\pm1.944\%$ for sub-0001 but only $0.069\pm0.012\%$ for sub-0002.
fMRI--Omni shows the same subject direction with substantially higher variance
on sub-0001 ($5.234\pm5.239\%$) and a near-floor mean for sub-0002
($0.050\pm0.037\%$).  The large subject difference, together with the
fMRI--Omni seed spread on sub-0001, motivates subject-wise reporting and
complete participant folds.

\FloatBarrier
\section{Training and Compute Details}

This section records hardware and precision; parameter counts and measured
training duration are reported in the main paper.

\subsection{Hardware and Software}

\begin{table}[!htbp]
    \centering
    \small
    \begin{tabular}{ll}
        \toprule
        Property     & Value                                         \\
        \midrule
        GPUs         & 2$\times$ NVIDIA A100 80GB PCIe               \\
        GPU memory   & 80 GB per GPU                                 \\
        CUDA         & 12.1                                          \\
        cuDNN        & 90100                                         \\
        PyTorch      & 2.5.1+cu121                                   \\
        Launch style & \texttt{accelerate launch -{}-num\_processes=2} \\
        \bottomrule
    \end{tabular}
    \caption{Hardware and software environment.}
    \label{tab:supp_hardware}
\end{table}

Archived execution metadata records bf16 for the reported Omni core runs and
fp16 for ImageBind startup.  The Omni artifacts record local batch 24 without
distributed embedding gathering; released YAML files provide current
reproduction defaults.


\end{document}